\documentclass{article}

\usepackage{graphicx} % more modern
\usepackage{subfigure} 

\usepackage{natbib}

\usepackage{algorithm}
\usepackage{algorithmic}

\usepackage{wrapfig}
\usepackage[rightcaption]{sidecap}

\usepackage{amsmath}
\usepackage{amssymb}
\usepackage{enumerate}
\DeclareMathOperator{\veco}{vec}   
\DeclareMathOperator*{\argmin}{arg\,min}
\usepackage{xcolor}

\newtheorem{de}{Definition}%[subsection] % les d�finitions et les th�or�mes sont
\newtheorem{theo}{Theorem}%[section]    % num�rot�s par section
\newtheorem{prop}[theo]{Proposition}    % Les propositions ont le m�me compteur
\usepackage{hyperref}

 \usepackage[accepted]{icml2013}

\icmltitlerunning{A Generalized Kernel Approach to Structured Output Learning}

\begin{document} 

\twocolumn[
\icmltitle{A Generalized Kernel Approach to Structured Output Learning}

% unless you've provided the [accepted] option to the icml2013
\icmlauthor{Hachem Kadri}{hachem.kadri@lif.univ-mrs.fr}
\icmladdress{Universit\'e d'Aix-Marseille, QARMA - LIF/CNRS, FRANCE}
\icmlauthor{Mohammad Ghavamzadeh}{mohammad.ghavamzadeh@inria.fr}
\icmladdress{INRIA Lille - Nord Europe, 
	      Team SequeL, FRANCE}
\icmlauthor{Philippe Preux}{philippe.preux@inria.fr}
\icmladdress{Universit\'e de Lille, LIFL/CNRS, INRIA, FRANCE}

\icmlkeywords{structured output learning, kernel dependency estimation, operator-valued kernel, covariance operator}

\vskip 0.3in
]

%%%%%%%%%%%%%%%%%%%%%%%%%%%%%%%%%%%%%%%%%%%%%%%%%%%%%%%%%%%%%%%%%%%%%%%%%%%%%%%%%%%%%%%%%%%%%%%%%%%%%%%%%%%%%%%%%%%%%%%%%%%%%%%%%%%%%%%%%%%%%%%%%%%%%%%%%%%%%%%%%%%%%%%%%%%%%%%%%%%%%%%%%%%%%%%%%%%%%%%%%%%%%%%%%%%%%%%%%%%%%%%%%%%%%%%%%%%%%%%%%%%%%%%%%%%%%%%%%%%%%%%%%%%%%%%%%%%%%%%%%%%%%

\begin{abstract} 
We study the problem of structured output learning from a regression perspective. 
We first provide a general formulation of the kernel dependency estimation~(KDE) approach to this problem using 
operator-valued kernels. Our formulation overcomes the two main limitations of the original KDE approach, 
namely the decoupling between outputs in the image space and the inability to use a joint feature space.
%We show that the existing KDE formulations are special cases of our framework. 
We then propose a covariance-based operator-valued kernel that allows us to take into account the structure of the 
kernel feature space. This kernel operates on the output space and only encodes the interactions between the outputs 
without any reference to the input space. To address this issue, we introduce a variant of our KDE method based on the 
conditional covariance operator that in addition to the correlation between the outputs takes into account the effects of 
the input variables. Finally, we evaluate the performance of our KDE 
approach %using both covariance and conditional covariance kernels
on three structured output problems, and compare it to the state-of-the-art kernel-based structured output regression 
methods.
\end{abstract} 

%%%%%%%%%%%%%%%%%%%%%%%%%%%%%%%%%%%%%%%%%%%%%%%%%%%%%%%%%%%%%%%%%%%%%%%%%%%%%%%%%%%%%%%%%%%%%%%%%%%%%%%%%%%%%%%%%%%%%%%%%%%%%%%%%%%%%%%%%%%%%%%%%%%%%%%%%%%%%%%%%%%%%%%%%%%%%%%%%%%%%%%%%%%%%%%%%%%%%%%%%%%%%%%%%%%%%%%%%%%%%%%%%%%%%%%%%%%%%%%%%%%%%%%%%%%%%%%%%%%%%%%%%%%%%%%%%%%%%%%%%%%%%

%\vspace{-0.3in}
\section{Introduction}
\label{sec:intro}
%\vspace{-0.05in}

In many practical problems such as statistical machine translation~\citep{Wang-2010} and speech recognition 
or synthesis~\citep{Cortes-KDE-2005}, we are faced with the task of learning a mapping between objects of 
different nature that each can be characterized by complex data structures.
%In many problems one is faced with the task of learning a mapping between objects of different nature, which can be generally characterized by complex data structures. This is often the case in a number of practical applications such as statistical machine translation~\cite{Wang-2010} and speech recognition or synthesis~\cite{Cortes-KDE-2005}. %ref and object localization~\cite{Blaschko-2008}. 
%; for example (natural language processing problems) ...
Therefore, designing algorithms that are sensitive enough to detect structural dependencies among these complex data is 
becoming increasingly important.~While classical learning algorithms can be easily extended to complex inputs, more refined 
and sophisticated algorithms are needed to handle complex outputs. In this case, several mathematical and methodological 
difficulties arise and these difficulties increase with the complexity of the output space. Complex output data can be 
divided into three classes: 
{\bf 1)}~{\em Euclidean}:~vectors or real-valued functions; 
{\bf 2)}~{\em mildly non-Euclidean}:~points on manifolds and shapes; and
{\bf 3)}~{\em strongly non-Euclidean}:~structured data like trees and graphs.~The focus in the machine learning and 
statistics communities has been mainly on multi-task learning (vector outputs) and functional data analysis (functional 
outputs)~\citep{Caruana-1997,Ramsay-2005}, %,Ando-2005
where in both cases output data reside in a Euclidean space, but there has also been considerable interest in expanding 
general learning algorithms to structured outputs. % data.

\begin{table*}[t]
%\begin{center}
\caption{This table summarizes the notations used in the paper.}
\centering
\vskip 0.1in
\begin{tabular}{|l|c||l|c|}
\hline
%name & notation & information & name & notation & information\\
%\hline 
input space & $\mathcal{X}$ & structured output space & $\mathcal{Y}$ \\
%number of samples & $n$ & & \\
input data & $x_i\in\mathcal{X}$ & structured output data & $y_i\in\mathcal{Y}$ \\
scalar-valued kernel & $k:\mathcal{X}\times\mathcal{X}\rightarrow\mathbb{R}$ & scalar-valued kernel  & $l:\mathcal{Y}\times\mathcal{Y}\rightarrow\mathbb{R}$ \\
output feature space & $\mathcal{F_Y}$ & output feature map & $\Phi_l:\mathcal{Y}\rightarrow\mathcal{F_Y}$ \\
% (a RKHS with kernel $l$) & & & \\
set of operators on $\mathcal{F_Y}$ to $\mathcal{F_Y}$ & $\mathcal{L(F_Y)}$ & operator-valued kernel  & $K:\mathcal{X}\times\mathcal{X}\rightarrow\mathcal{L(F_Y)}$ \\
joint feature space  & $\mathcal{F_{XY}}$ & joint feature map & $\Phi_K:\mathcal{X}\times\mathcal{F_Y}\rightarrow\mathcal{F_{XY}}$ \\
%(a RKHS with kernel $K$) & & & \\
a mapping from $\mathcal{X}$ to $\mathcal{Y}$ & $f$ & a mapping from $\mathcal{X}$ to $\mathcal{F_Y}$ & $g$ \\
\hline
\end{tabular}
\label{tab:notation}
%\vspace{-0.15in}
\end{table*}

One difficulty encountered when working with structured data is that usual Euclidean methodology cannot be applied in 
this case. Reproducing kernels provide an elegant way to overcome this problem. Defining a suitable kernel on the 
structured data allows to encapsulate the structural information in a kernel function and transform the problem to a 
Euclidean space. Two different, but closely related, kernel-based approaches for structured output learning can be found 
in the literature~\citep{Bakir-2007}: {\em kernel dependency estimation}~(KDE) and {\em joint kernel maps}~(JKM). KDE is 
a regression-based approach that was first proposed by~\citet{Weston-KDE-2003} and then reformulated 
by~\citet{Cortes-KDE-2005}.
The idea is to define a kernel on the output space $\mathcal{Y}$ to project the structured output to a 
real-valued reproducing kernel Hilbert space~(RKHS) $\mathcal{F_Y}$, and then perform a {\em scalar-valued} kernel 
ridge regression~(KRR) between the input space $\mathcal{X}$ and the feature space $\mathcal{F_Y}$. Having the 
regression coefficients, the prediction is obtained by computing the pre-image from $\mathcal{F_Y}$. 
On the other hand, the JKM approach is based on joint kernels, which are nonlinear similarity measures between 
input-output pairs~\citep{Tsochantaridis-2005, Weston-JKM-2007}. While in KDE separate kernels are used to project 
input and output data to two (possibly different) feature spaces, the joint kernel in JKM maps them into a single 
feature space, which then allows us to take advantage of our prior knowledge on both input-output and output correlations. 
However, this improvement requires an exhaustive pre-image computation during training, a problem that is encountered by 
KDE only in the test phase. Avoiding this computation during training is an important advantage of KDE over JKM methods.

In this paper, we focus on the KDE approach to structured output learning.~The main contributions of this paper can be 
summarized as follows:~{\bf 1)} Building on the works of~\citet{Caponnetto-2007} and~\citet{Flo-2011}, we propose a more general KDE 
formulation (prediction and pre-image steps) based on {\em operator-valued} (multi-task) 
kernels instead of {\em scalar-valued} ones used by the existing methods~(Sec.~\ref{sec:kde}).~This extension allows KDE to capture the dependencies between the outputs as well as between the input and output 
variables, which is an improvement over the existing KDE methods that fail to take into account these 
dependencies. 
{\bf 2)} We also propose a variant (generalization) of the kernel trick to cope with the technical difficulties 
encountered when working with operator-valued kernels (Sec.~\ref{sec:kde}). This allows us to {\bf (i)} formulate 
the pre-image problem using only kernel functions (not feature maps that cannot be computed explicitly), and {\bf (ii)} 
avoid the computation of the inner product between feature maps after being modified with an operator whose role is to capture the structure of complex objects.~{\bf 3)}~We then introduce a novel family of operator-valued kernels, based on covariance operators on RKHSs, that 
allows us to take full advantage of our KDE formulation.~These kernels offer a simple and powerful way to address 
the main limitations of the original KDE formulation, namely the decoupling between outputs in the image space and the 
inability to use a joint feature space~(Sec.~\ref{sec:cov}). {\bf 4)}~We show how the pre-image problem, in the case of 
covariance and conditional covariance operator-valued kernels, can be expressed only in terms of input and output Gram 
matrices, and provide a low rank approximation to efficiently compute it (Sec.~\ref{sec:cov}).~{\bf 5)} Finally, we 
empirically evaluate the performance of our proposed KDE approach and show its effectiveness on three structured output 
prediction problems involving numeric and non-numerical outputs~(Sec.~\ref{sec:exp}).
%%% 
%%%
%%%
It should be noted that generalizing KDE using operator-valued kernels was first proposed in~\citet{Flo-2011}.
The authors have applied this generalization to the problem of link prediction which did not require a pre-image step.  Based on this work, we discuss both the regression and pre-image steps of operator-valued KDE, propose new covariance-based operator-valued kernels and show how they can be implemented efficiently.

%%%%%%%%%%%%%%%%%%%%%%%%%%%%%%%%%%%%%%%%%%%%%%%%%%%%%%%%%%%%%%%%%%%%%%%%%%%%%%%%%%%%%%%%%%%%%%%%%%%%%%%%%%%%%%%%%%%%%%%%%%%%%%%%%%%%%%%%%%%%%%%%%%%%%%%%%%%%%%%%%%%%%%%%%%%%%%%%%%%%%%%%%%%%%%%%%%%%%%%%%%%%%%%%%%%%%%%%%%%%%%%%%%%%%%%%%%%%%%%%%%%%%%%%%%%%%%%%%%%%%%%%%%%%%%%%%%%%%%%%%%%%%

%\vspace{-0.1in}
\section{Preliminaries}
\label{sec:back}
%\vspace{-0.05in}

In this section, we first introduce the notations used throughout the paper, lay out the setting of the problem studied 
in the paper, and provide a high-level description of our approach to this problem. Then before reporting our 
KDE formulation, we provide a brief overview of operator-valued kernels and their associated RKHSs. To assist 
the reading, we list the notations used in the paper in Table~\ref{tab:notation}.

%%%%%%%%%%%%%%%%%%%%%%%%%%%%%%%%%%%%%%%%%%%%%%%%%%%%%%%%%
%%%%%%%%%%%%%%%%%%%%%%%%%%%%%%%%%%%%%%%%%%%%%%%%%%%%%%%%%
%%%%%%%%%%%%%%%%%%%%%%%%%%%%%%%%%%%%%%%%%%%%%%%%%%%%%%%%%

%\vspace{-0.05in}
\subsection{Problem Setting and Notations}
\label{subsec:prob}

%To assist the reading, we list the notations used in the paper in Table~\ref{tab:notation}.

\begin{figure}[t]
\centering
\includegraphics[width=0.75\columnwidth]{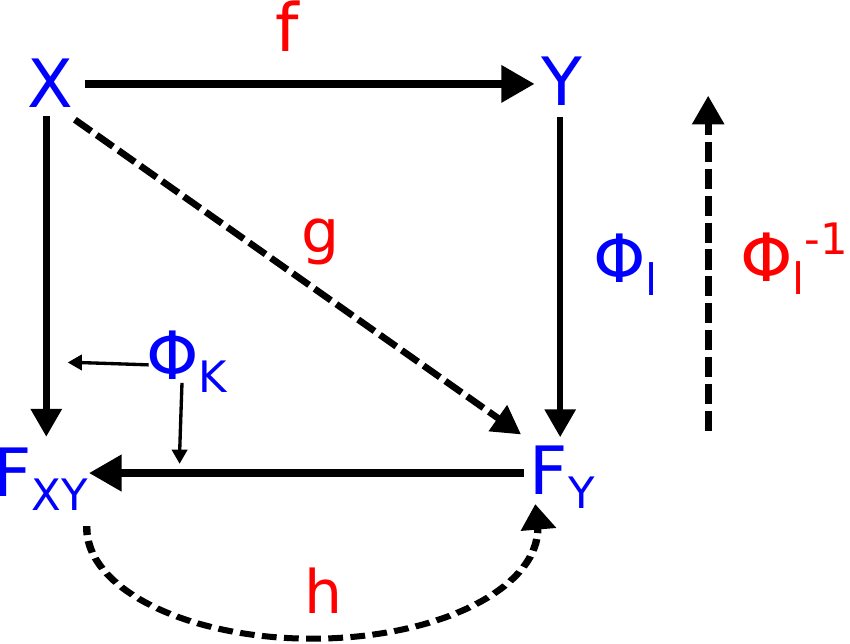}
%\vspace{-0.1in}
\caption{Kernel Dependency Estimation. Our generalized formulation consists of learning the mapping $g$ using an operator-valued kernel ridge regression rather than a scalar-valued one as in the formulations of~\citet{Weston-KDE-2003} and~\citet{Cortes-KDE-2005}. Using an operator-valued kernel mapping, we construct a joint feature space from information of input and output spaces in which input-output and output correlations can be taken into account.}
%\vspace{-0.2in}
\label{fig:kde2}
\end{figure}

Given $(x_i,y_i)_{i=1}^n \in \mathcal{X} \times \mathcal{Y}$, where $\mathcal{X}$ and $\mathcal{Y}$ are the input 
and structured output spaces, we consider the problem of learning a mapping $f$ from $\mathcal{X}$ to $\mathcal{Y}$. 
The idea of KDE is to embed the output data using a mapping $\Phi_l$ between the structured output space $\mathcal{Y}$ 
and a Euclidean feature space $\mathcal{F_Y}$ defined by a scalar-valued kernel $l$. Instead of learning $f$ in order 
to predict an output $y$ for an input $x$, the KDE methods first learn the mapping $g$ from $\mathcal{X}$ to 
$\mathcal{F_Y}$, and then compute the pre-image of $g(x)$ by the inverse 
mapping of $\Phi_l$, i.e.,~$y = f(x) = \Phi_l^{-1}\big(g(x)\big)$ (see Fig.~\ref{fig:kde2}). 
%
%Note that in the original KDE algorithm~\cite{Weston-KDE-2003}, the feature space $\mathcal{F_Y}$ is reduced using kernel PCA. 
All existing KDE methods use ridge regression with a scalar-valued kernel $k$ on $\mathcal{X} \times \mathcal{X}$ to learn the 
mapping~$g$. This approach has the drawback of not taking into account the dependencies between the data in the 
feature space~$\mathcal{F_Y}$. The variables in $\mathcal{F_Y}$ can be highly correlated, since they are the 
projection of $y_i$'s using the mapping $\Phi_l$, and taking this correlation into account is essential to retain and 
exploit the structure of the outputs. To overcome this problem, our KDE approach uses an operator-valued  (multi-task) 
kernel to encode the relationship between the output components, and learns $g$ in the 
vector-valued (function-valued) RKHS built by this kernel using operator-valued kernel-based regression. 
The advantage of our formulation is that it allows vector-valued regression to be performed by directly optimizing 
over a Hilbert space of vector-valued functions, instead of solving independent scalar-valued regressions. 
As shown in Fig.~\ref{fig:kde2}, the feature space induced by an operator-valued kernel, $\mathcal{F_{XY}}$, 
is a joint feature space that contains information of both input space $\mathcal{X}$ and output feature 
space $\mathcal{F_Y}$. This allows KDE to exploit both the output and input-output correlations. 
The operator-valued kernel implicitly induces a metric on the joint space of inputs and output features, 
and then provides a powerful way to devise suitable metrics on the output features, which can be changed 
depending on the inputs. In this sense, our proposed method is a natural way to incorporate prior knowledge about 
the structure of the output feature space while taking into account the inputs.

%%%%%%%%%%%%%%%%%%%%%%%%%%%%%%%%%%%%%%%%%%%%%%%%%%%%%%%%%%%%
%%%%%%%%%%%%%%%%%%%%%%%%%%%%%%%%%%%%%%%%%%%%%%%%%%%%%%%%%%%%

%%%%%%%%%%%%%%%%%%%%%%%%%%%%%%%%%%%%%%%%%%%%%%%%%%%%%%%%%%%%
%%%%%%%%%%%%%%%%%%%%%%%%%%%%%%%%%%%%%%%%%%%%%%%%%%%%%%%%%%%%

%\vspace{-0.05in}
\subsection{Operator-valued Kernels and Associated RKHSs}

We now provide a few definitions related to operator-valued kernels and their associated RKHSs that are used in the paper (see~\citep{Micchelli-2005a, Caponnetto-2008, Alvarez-2012} for more details). % , Carmeli-2010
These kernel spaces have recently received more attention, since they are suitable for leaning in problems where the outputs are vectors (as in multi-task learning~\citep{Evgeniou-2005}) or functions (as in functional regression~\citep{Kadri-2010}) instead of scalars.
Also, it has been shown recently that these spaces are appropriate for learning conditional mean embeddings~\citep{Grunewalder-2011a}.
Let $\mathcal{L(F_Y)}$ be the set of bounded operators from $\mathcal{F_Y}$ to $\mathcal{F_Y}$. 
\begin{de}(Non-negative $\mathcal{L(F_Y)}$-valued kernel) A non-negative $\mathcal{L(F_Y)}$-valued kernel $K$  is an operator-valued function on $\mathcal{X} \times \mathcal{X}$, i.e.,~$K:\mathcal{X} \times \mathcal{X}\rightarrow\mathcal{L(F_Y)}$, such that: 
%\vspace{-0.2cm}
\begin{enumerate}[i.]
\item $\forall x_i, x_j \in \mathcal{X},\;K(x_i,x_j) = K(x_j,x_i)^*$ ($*$ denotes the adjoint),
\item $\forall m\in\mathbb{N}_+^*,\; \forall x_1,\ldots,x_m \in \mathcal{X},\;\forall \varphi_i, \varphi_j\in \mathcal{F_Y} \\ \sum\limits_{i,j=1}^m \langle K(x_i,x_j) \varphi_j, \varphi_i\rangle_{\mathcal{F_Y}} \geq 0$.
\end{enumerate}
\end{de}
%
%\vspace{-0.25cm}
The above properties guarantee that the operator-valued kernel matrix $\mathbf{K}=\big[K(x_i,x_j)\in\mathcal{L(F_Y)}\big]_{i,j=1}^n$ is positive definite. Given a non-negative $\mathcal{L(F_Y)}$-valued kernel $K$ on $\mathcal{X}\times\mathcal{X}$, there exists a unique RKHS of $\mathcal{F_Y}$-valued functions whose reproducing kernel is $K$. 
%
%\vspace{-0.35cm}
\begin{de}($\mathcal{F_Y}$-valued RKHS) A RKHS $\mathcal{F_{XY}}$ of $\mathcal{F_Y}$-valued functions $g : \mathcal{X} \rightarrow \mathcal{F_Y}$ is a Hilbert space such that there is a non-negative $\mathcal{L(F_Y)}$-valued kernel $K$ with the following properties: 
\vspace{-0.2cm}
\begin{enumerate}[i.]
\item \label{enum1:i} $\forall x\in\mathcal{X},\;\forall\varphi\in\mathcal{F_Y}\hspace{0.5in} K(x,\cdot)\varphi\in\mathcal{F_{XY}}$, 
%\item \label{enum1:ii} $\forall g\in\mathcal{F_{XY}},\quad\quad\langle g,K(x,\cdot)\varphi\rangle _{\mathcal{F_{XY}}} = \langle g(x),\varphi\rangle _{\mathcal{F_Y}}$. %\hspace*{0.1cm} (reproducing property).
\item \label{enum1:ii} $\forall g\in\mathcal{F_{XY}},\;\forall x\in\mathcal{X},\;\forall\varphi\in\mathcal{F_Y} \\ \langle g,K(x,\cdot)\varphi\rangle _{\mathcal{F_{XY}}} = \langle g(x),\varphi\rangle _{\mathcal{F_Y}}$.
\end{enumerate}
\end{de}
Every RKHS $\mathcal{F_{XY}}$ of $\mathcal{F_Y}$-valued functions is associated with 
a unique non-negative $\mathcal{L(F_Y)}$-valued kernel $K$, called the reproducing kernel.  
%
%For further reading on operator-valued kernels and their associated RKHSs we recommend~\cite{Micchelli-2005a, Caponnetto-2008, Carmeli-2010, Alvarez-2011}.

%%%%%%%%%%%%%%%%%%%%%%%%%%%%%%%%%%%%%%%%%%%%%%%%%%%%%%%%%%%%
%%%%%%%%%%%%%%%%%%%%%%%%%%%%%%%%%%%%%%%%%%%%%%%%%%%%%%%%%%%%
%%%%%%%%%%%%%%%%%%%%%%%%%%%%%%%%%%%%%%%%%%%%%%%%%%%%%%%%%%%%

%\vspace{-0.1in}
\section{Operator-valued Kernel Formulation of Kernel Dependency Estimation}
\label{sec:kde}
%\vspace{-0.05in}

In this section, we describe our operator-valued KDE formulation in which the feature spaces associated to input and output kernels can be infinite dimensional, contrary to~\citet{Cortes-KDE-2005} that only considers finite feature spaces. Operator-valued KDE is performed in two steps: 

%\vspace{-0.15in}
\noindent
\textbf{Step~1 (kernel ridge) Regression:} We use operator-valued kernel-based regression and learn the function $g$ in the $\mathcal{F_Y}$-valued RKHS $\mathcal{F_{XY}}$ from the training data $\big(x_i,\Phi_l(y_i)\big)_{i=1}^n \in \mathcal{X}\times\mathcal{F_Y}$, where $\Phi_l$ is the mapping from the structured output space $\mathcal{Y}$ to the scalar-valued RKHS $\mathcal{F_Y}$. Similar to other KDE formulations, we consider the following regression problem: %kernel ridge regression problem % of the form: %and solve the optimization problem
\begin{equation}
\label{eq:rr}
 \argmin\limits_{g \in \mathcal{F_{XY}}} \sum\limits_{i=1}^n \| g(x_i) - \Phi_l(y_i) \|_{\mathcal{F_Y}}^2 + \lambda \|g\|^2,
\end{equation}
where $\lambda > 0$ is a regularization parameter. Using the representer theorem in the vector-valued setting~\citep{Micchelli-2005a}, 
the solution of~\eqref{eq:rr} has the following form\footnote{As in the scalar-valued case, operator-valued kernels provide an elegant way of dealing with nonlinear problems~(mapping $g$) by reducing them to linear ones~(mapping $h$) in some feature space~$\mathcal{F_{XY}}$ (see Figure~\ref{fig:kde2}).}:
\begin{equation}
\label{eq:rt}
 g(\cdot) = \sum\limits_{i=1}^n K(\cdot,x_i) \psi_i\;,
\end{equation}
where $\psi_i \in \mathcal{F_Y}$. Using~\eqref{eq:rt}, we obtain an analytic solution for the optimization problem~\eqref{eq:rr} as
\begin{equation}
\label{eq:rrsol}
\mathbf{\Psi} = (\mathbf{K} + \lambda I)^{-1} \mathbf{\Phi_l}\;,
\end{equation}
where $\mathbf{\Phi_l}$ is the column vector of $\big[\Phi_l(y_i)\in\mathcal{F_Y}\big]_{i=1}^n$. Eq.~\ref{eq:rrsol} 
is a generalization of the scalar-valued kernel ridge regression solution to vector or functional 
outputs~\citep{Caponnetto-2007}, in which the kernel matrix is a block operator matrix.  %ref  , Kadri-ICML-2011
Note that features $\Phi_l(y_i)$ in this equation can be explicit or implicit. We show in the following that even with implicit features, we are able to formulate the structured output prediction problem in terms of explicit quantities that are computable via input and output kernels.
% note that at this stage an analytical solution in terms of the implicit features is derived in "Step 1", which is later substituted into "Step 2"
% 

\noindent
\textbf{Step 2 (pre-image) Prediction:} In order to compute the structured prediction $f(x)$ for an input $x$, we solve the following pre-image problem:
\begin{align*}
f(x) & = \argmin\limits_{y\in \mathcal{Y}} \| g(x) - \Phi_l(y) \|_{\mathcal{F_Y}}^2 \\
&= \argmin\limits_{y\in \mathcal{Y}} \| \sum\limits_{i=1}^n K(x_i,x)\psi_i - \Phi_l(y) \|_{\mathcal{F_Y}}^2 \\
& = \argmin\limits_{y\in \mathcal{Y}} \| \mathbf{K}_x \mathbf{\Psi} - \Phi_l(y) \|_{\mathcal{F_Y}}^2 \\ 
& = \argmin\limits_{y\in \mathcal{Y}} \| \mathbf{K}_x (\mathbf{K} + \lambda I)^{-1} \mathbf{\Phi_l} - \Phi_l(y) \|_{\mathcal{F_Y}}^2 \\
& = \argmin\limits_{y\in \mathcal{Y}}\;l(y,y) - 2\langle  \mathbf{K}_x (\mathbf{K} + \lambda I)^{-1} \mathbf{\Phi_l}, \Phi_l(y)\rangle_{\mathcal{F_Y}}
\end{align*}
where $\mathbf{K}_x$ is the row vector of operators corresponding to input $x$. In many problems, the kernel map $\Phi_l$ is unknown and only implicitly defined through the kernel $l$. In these problems, the above operator-valued kernel formulation has an inherent difficulty in expressing the pre-image problem and the usual kernel trick is not sufficient to solve it. To overcome this problem, we introduce the following variant (generalization) of the kernel trick: $\langle \mathcal{T} \Phi_l(y_1),\Phi_l(y_2)\rangle_\mathcal{F_Y} = [\mathcal{T}l(y_1,\cdot)](y_2)$, where $\mathcal{T}$ is an operator in $\mathcal{L(F_Y)}$. Note that the usual kernel trick $\langle \Phi_l(y_1),\Phi_l(y_2)\rangle_\mathcal{F_Y} = l(y_1,y_2)$ is recovered from this variant when $\mathcal{T}$ is the identity operator. It is easy to check that our proposed trick holds if we consider the feature space associated to the kernel $l$, i.e.,~$\Phi_l(y) = l(y,.)$. A proof for the more general case in which the features $\Phi_l$ can be any 
implicit mapping of a Mercer kernel is given for self-adjoint operator $\mathcal{T}$ %in Appendix~A 
in~\citep[Appendix~A]{kadri-2013-ICML_TechRep}. Using this trick,  we may now express $f(x)$ using only kernel functions: %the pre-image problem may be rewritten as
\begin{equation}
\label{eq:preim}
 f(x) = \argmin\limits_{y\in \mathcal{Y}}\;l(y,y) - 2 \big[\mathbf{K}_x (\mathbf{K} + \lambda I)^{-1} \mathbf{L}_\bullet\big](y),
%\sum\limits_i [(K_x (\mathbf{K} + \lambda I)^{-1})_i l(y_i,.)](y)
\end{equation}
where $\mathbf{L}_\bullet$ is the column vector whose $i$'th component is $l(y_i,\cdot)$. 
%We may now express $f(x)$ only using kernel functions. 
%
%
Note that the KDE regression and prediction steps of~\citet{Cortes-KDE-2005} can be recovered from Eqs.~\ref{eq:rrsol} and~\ref{eq:preim} using an operator-valued kernel $K$ of the form $K(x_i,x_j) = k(x_i,x_j)\mathcal{I}$, in which $k$ is a scalar-valued kernel and $\mathcal{I}$ is the identity operator in $\mathcal{F_Y}$.
%
% % 
% % Note that the regression and prediction steps of the KDE formulation in~\citet{Cortes-KDE-2005} can be recovered from Eqs.~\ref{eq:rrsol} and~\ref{eq:preim} using an operator-valued kernel $K$ of the form $K(x_i,x_j) = k(x_i,x_j)\mathcal{I}$, in which $k$ is a scalar-valued kernel on $\mathcal{X}$ and $\mathcal{I}$ is the identity operator in $\mathcal{F_Y}$. 

Now that we have a general formulation of KDE, we turn our attention to build operator-valued kernels that can take into account the structure of the kernel feature space~$\mathcal{F_Y}$ as well as input-output and output correlations. This is described in the next section.
 
%%%%%%%%%%%%%%%%%%%%%%%%%%%%%%%%%%%%%%%%%%%%%%%%%%%%%%%%%
%%%%%%%%%%%%%%%%%%%%%%%%%%%%%%%%%%%%%%%%%%%%%%%%%%%%%%%%%
%%%%%%%%%%%%%%%%%%%%%%%%%%%%%%%%%%%%%%%%%%%%%%%%%%%%%%%%%
%%%%%%%%%%%%%%%%%%%%%%%%%%%%%%%%%%%%%%%%%%%%%%%%%%%%%%%%%
%%%%%%%%%%%%%%%%%%%%%%%%%%%%%%%%%%%%%%%%%%%%%%%%%%%%%%%%%

%\vspace{-0.1in}
\section{Covariance-based Operator-valued Kernels}
\label{sec:cov}
%\vspace{-0.05in}

In this section, we study the problem of designing operator-valued kernels suitable for structured outputs in the KDE formulation. This is quite important in order to take full advantage of the operator-valued KDE formulation. The main purpose of using the operator-valued kernel formulation is to take into account the dependencies between the variables $\Phi_l(y_i)$, i.e.,~the projection of $y_i$ in the feature space $\mathcal{F_Y}$, with the objective of capturing the structure of the output data encapsulated in $\Phi_l(y_i)$. Operator-valued kernels have been studied more in the context of multi-task learning, where the output is assumed to be in $\mathbb{R}^d$ with $d$ the number of tasks~\citep{Evgeniou-2005}. %,Caponnetto-2008
Some work has also been focused on extending these kernels to the domain of functional data analysis to deal with the problem of regression with functional responses, where outputs are considered to be in the $L^2$-space~\citep{Kadri-2010}. %Lian-2007, 
However, the 
operator-valued Kernels used in these contexts for discrete~(vector) or continuous~(functional) outputs cannot be used in our formulation. This is because in our case the feature space $\mathcal{F_Y}$ can be known only implicitly by the output kernel~$l$, and depending on $l$, $\mathcal{F_Y}$ can be finite or infinite dimensional. Therefore, we focus our attention to operators that act on scalar-valued RKHSs. Covariance operators on RKHS have recently received considerable attention. %amount of attention in the machine learning community. 
These operators that provide the simplest measure of dependency have been successfully applied to the problem of dimensionality reduction~\citep{Fukumizu-2004}, and played an important role in dealing with a number of statistical test problems~\citep{Gretton-2005}. We use the following covariance-based operator-valued kernel in our KDE formulation:  %ref ,Fukumizu-2008
\begin{equation}
\label{eq:kern1}
 K(x_i,x_j) = k(x_i,x_j) C_{YY},
\end{equation}
where $k$ is a scalar-valued kernel and $C_{YY}: \mathcal{F_Y} \rightarrow \mathcal{F_Y}$ is the covariance operator defined for a random~variable $Y$ on $\mathcal{Y}$ as $\langle \varphi_i, C_{YY} \varphi_j \rangle_{\mathcal{F_Y}} = \mathbb{E}\big[\varphi_i(Y)\varphi_j(Y)\big]$. The empirical covariance operator $\widehat{C}_{YY}^{(n)}$ is given by
\begin{equation}
\widehat{C}_{YY}^{(n)}=\frac{1}{n}\sum\limits_{i=1}^n l(\cdot,y_i)\otimes l(\cdot,y_i),
\end{equation}
where $\otimes$ is the tensor product $(\varphi_1\otimes \varphi_2)h = \langle \varphi_2,h \rangle \varphi_1$. The operator-valued kernel~(\ref{eq:kern1}) is nonnegative since  
%
%\vspace{-0.225in}
%\begin{small}
\begin{align*}
&\sum\limits_{i,j=1}^m \langle K(x_i,x_j) \varphi_j, \varphi_i\rangle_{\mathcal{F_Y}} = \sum\limits_{i,j}^m \langle k(x_i,x_j) \widehat{C}_{YY}^{(n)}\varphi_j, \varphi_i\rangle \\
&= \sum\limits_{p=1}^n \sum\limits_{i,j}^m \frac{1}{n} \langle l(.,y_p), \varphi_i \rangle k(x_i,x_j) \langle l(.,y_p), \varphi_j \rangle \geq 0.
\end{align*}
%\end{small}
%\vspace{-0.175in}
%
The last step is due to the positive definiteness of $k$.% semi-definiteness of the scalar-valued kernel $k$. 

The kernel~\eqref{eq:kern1} is a separable operator-valued kernel since it operates on the output space, and then encodes the interactions between the outputs, without any reference to the input space. Although this property can be restrictive in specifying input-output correlations, because of its simplicity, most of the operator-valued kernels proposed in the literature belong to this category (see~\citep{Alvarez-2012} for a review of separable and beyond separable operator-valued kernels). To address this issue, we propose a variant of the kernel in~\eqref{eq:kern1} based on the conditional covariance operator, %, rather than the covariance operator, i.e.,
\begin{equation}
 \label{eq:kern2}
 K(x_i,x_j) = k(x_i,x_j) C_{YY|X},
\end{equation}
where $C_{YY|X} = C_{YY} - C_{YX}C_{XX}^{-1}C_{XY}$ is the conditional covariance operator on $\mathcal{F_Y}$. This operator allows the operator-valued kernel to simultaneously encode the correlations between the outputs and to take into account (non-parametrically) the effects of the inputs. In Proposition~\ref{prop:cov}, we show how the pre-image problem~(\ref{eq:preim}) can be formulated using the covariance-based operator-valued kernels in~\eqref{eq:kern1} and~\eqref{eq:kern2}, and expressed in terms of input and output Gram matrices. The proof is reported % in Appendix B 
in~\citep[Appendix~B]{kadri-2013-ICML_TechRep}.

\begin{prop}
\label{prop:cov}
The pre-image problem of Eq.~\ref{eq:preim} can be written for covariance and conditional covariance operator-valued kernels defined by Eqs.~\eqref{eq:kern1} and~\eqref{eq:kern2} as 
\begin{equation}
\label{eq:prop1}
\argmin\limits_{y\in \mathcal{Y}}\;l(y,y) - 2 \mathbf{L}_y^\top (\mathbf{k}_x^\top \otimes \mathbf{T}) (\mathbf{k} \otimes \mathbf{T} + n \lambda I_{n^2})^{-1} \veco(I_{n}),
\end{equation}
where $\mathbf{T} = \mathbf{L}$ for the covariance operator and $\mathbf{T} = \mathbf{L} - (\mathbf{k} + n \epsilon I_n)^{-1}\mathbf{k} \mathbf{L}$ for the conditional covariance operator in which $\epsilon$ is a regularization parameter required for the operator inversion, $\mathbf{k}$ and $\mathbf{L}$ are Gram matrices associated to the scalar-valued kernels $k$ and $l$, $\mathbf{k}_x$ and $\mathbf{L}_y$ are the column vectors $\big(k(x,x_1),\ldots, k(x,x_n)\big)^\top$ and $\big(l(y,y_1),\ldots, l(y,y_n)\big)^\top$, $\veco$ is the vector operator such that $\veco(A)$ is the vector of columns of the matrix $A$, and finally $\otimes$ is the Kronecker product.  
%
% Using the conditional covariance operator-valued kernel~(\eqref{eq:kern2}), the Gram matrix expression
% of the pre-image is obtained by replacing the kernel matrix $\mathbf{L}$ by
% $\mathbf{L} - (\mathbf{k} + n \epsilon I_n)^{-1}\mathbf{k} \mathbf{L}$, where $\epsilon$
% is a regularization parameter.
% %
\end{prop}

Note that in order to compute Eq.~\ref{eq:prop1} we need to store and invert the $n^2\times n^2$ matrix $(\mathbf{k} \otimes \mathbf{T} + n \lambda I_{n^2})$, which leads to space and computational complexities of order $O(n^4)$ and $O(n^6)$, respectively. However, we show %in Appendix C in the supplementary material 
that this computation can be performed more efficiently with space and computational complexities of order $O\big(\max(nm_1m_2,m_1^2m_2^2)\big)$ and $O(m_1^3m_2^3)$ using incomplete Cholesky decomposition~\citep{Bach-2002}, where generally $m_1 \ll n$ and $m_2 \ll n$; see~\citep[Appendix~C]{kadri-2013-ICML_TechRep} for more details. 

%%%%%%%%%%%%%%%%%%%%%%%%%%%%%%%%%%%%%%%%%%%%%%%%%%%%%%%%%
%%%%%%%%%%%%%%%%%%%%%%%%%%%%%%%%%%%%%%%%%%%%%%%%%%%%%%%%%
%%%%%%%%%%%%%%%%%%%%%%%%%%%%%%%%%%%%%%%%%%%%%%%%%%%%%%%%%
%%%%%%%%%%%%%%%%%%%%%%%%%%%%%%%%%%%%%%%%%%%%%%%%%%%%%%%%%
%%%%%%%%%%%%%%%%%%%%%%%%%%%%%%%%%%%%%%%%%%%%%%%%%%%%%%%%%

%\vspace{-0.1in}
\section{Related Work}
\label{sec:rw}
%\vspace{-0.05in}

In this section, we discuss related work on kernel-based structured output learning and compare it with our proposed operator-valued kernel formulation. 
%
% % In this section, we provide a brief overview of the related work and compare them with our proposed operator-valued kernel formulation. 
%This comparison indicates that our formulation is an important step towards a general kernelized framework to deal with complex input-output problems.  

%%%%%%%%%%%%%%%%%%%%%%%%%%%%%%%%%%%%%%%%%%%%%%%%%%%%%%%%
%%%%%%%%%%%%%%%%%%%%%%%%%%%%%%%%%%%%%%%%%%%%%%%%%%%%%%%%

%\vspace{-0.05in}
\subsection{KDE}
%\vspace{-0.05in} 

The main goal of our operator-valued KDE is to generalize KDE by taking into account input-output and output correlations. Existing KDE formulations try to address this issue either by performing a kernel PCA to decorrelate the outputs~\citep{Weston-KDE-2003}, or by incorporating some form of prior knowledge in the regression step using some specific constraints on the regression matrix which performs the mapping between input and output feature spaces~\citep{Cortes-KDE-2007}. 
%%%%%%%%%%%%%%%%%%%%%%%%%%%%%%%%
%%%%%%%%%%%%%%%%%%%%%%%%%%%%%%
Compared to kernel PCA~{\bf 1)}~our KDE formulation does not need to  have a dimensionality reduction step, which may cause loss of information when the spectrum of the output kernel matrix does not decrease rapidly,~{\bf 2)}~it does not require to assume that the dimensionality of the low-dimensional subspace (the number of principal components) is known and fixed in advance, and more importantly~{\bf 3)}~it succeeds to take into account the effect of the explanatory variables (input data). %when uses the conditional covariance operator-valued kernel. % (contrary to the existing KDE formulations)
Moreover, in contrast to~\citep{Cortes-KDE-2007}, our approach allows us to deal with infinite-dimensional feature spaces, 
  and encodes prior knowledge on input-output dependencies without requiring any particular form of constraints between input and output mappings. Indeed, information about the output space can be taken into account by the output kernel, and then the conditional covariance operator-valued kernel is a natural way to capture this information and also input-output relationships, independently of the dimension of the output feature space. 

%%%%%%%%%%%%%%%%%%%%%%%%%%%%%%%%%%%%%%%%%%%%%%%%%%%%%%%%%
%%%%%%%%%%%%%%%%%%%%%%%%%%%%%%%%%%%%%%%%%%%%%%%%%%%%%%%%%
%%%%%%%%%%%%%%%%%%%%%%%%%%%%%%%%%%%%%%%%%%%%%%%%%%%%%%%%%

%\vspace{-0.05in}
\subsection{Joint Kernels Meet Operator-valued Kernels}
%\vspace{-0.05in} 

Another approach to take into account input-output correlations is to use joint kernels, that are scalar-valued functions~(similarity measure) of input-output pairs~\citep{Weston-JKM-2007}.~In this context, the problem of learning the mapping $f$ from $\mathcal{X}$ to $\mathcal{Y}$ is reformulated as learning a function $\hat{f}$ from $\mathcal{X} \times \mathcal{Y}$ to $\mathbb{R}$ using a joint kernel~(JK)~\cite{Tsochantaridis-2005}.  
Our operator-valued kernel formulation includes the JK approach.~Similar to joint kernels, operator-valued kernels induce (implicitly) a similarity measure between input-output pairs. %~\citep{Micchelli-2005a,Evgeniou-2005}. 
This can be seen from the feature space formulation of operator-valued kernels~\citep{Caponnetto-2008,Kadri-ICML-2011}.~A feature map associated with an operator-valued kernel $K$ is a continuous function $\Phi_K$ such that
$ \langle K(x_1,x_2)\Phi_l(y_1),  \Phi_l(y_2)\rangle  =  \langle \Phi_K\big(x_1,\Phi_l(y_1)\big),\Phi_K\big(x_2,\Phi_l(y_2)\big) \rangle$.~So, the joint kernel is an inner product between an output $\Phi_l(y_2)$ and the result of applying the operator-valued kernel $K$ to another output $\Phi_l(y_1)$. We now show how two joint kernels in the literature~\cite{Weston-JKM-2007} can be recovered by a suitable choice of operator-valued kernel. 
\\[0.1cm]
{\bf 1)} {\em Tensor product JK:} $J\big((x_1,y_1),(x_2,y_2)\big) = k(x_1,x_2) l(y_1,y_2)\;$ can be recovered from the operator-valued kernel $K(x_1,x_2) = k(x_1,x_2)\mathcal{I}$, where $\mathcal{I}$ is the identity operator in $\mathcal{F_Y}$.
\\[0.1cm]
{\bf 2)} {\em Diagonal regularization JK:} $J\big((x_1,y_1),(x_2,y_2)\big) = (1-\lambda) k(x_1,x_2)\langle y_1,y_2 \rangle + \lambda \sum\limits_{i=1}^q x_1^i x_2^i y_1^i y_2^i\;$ can be recovered by selecting the output kernel $l(y_1,y_2)=\langle y_1,y_2 \rangle$ and the operator-valued kernel $K(x_1,x_2) = \big[(1-\lambda) k(x_1,x_2)\big]\mathcal{I} + \lambda \odot_{x_1\odot x_2}$, where $\odot$ is the point-wise product operator.%, 
%%%%%%%%%%%%%%%%%%%%%%%%%%%%%%%%%%%%%%%%%%%%%%%%%%%%%%%%%
%%%%%%%%%%%%%%%%%%%%%%%%%%%%%%%%%%%%%%%%%%%%%%%%%%%%%%%%%
%%%%%%%%%%%%%%%%%%%%%%%%%%%%%%%%%%%%%%%%%%%%%%%%%%%%%%%%%
%%%%%%%%%%%%%%%%%%%%%%%%%%%%%%%%%%%%%%%%%%%%%%%%%%%%%%%%%
%%%%%%%%%%%%%%%%%%%%%%%%%%%%%%%%%%%%%%%%%%%%%%%%%%%%%%%%%

%\vspace{-0.125in}
\section{Experimental Results}
\label{sec:exp}
%\vspace{-0.05in} 

%In this section, 
We evaluate our operator-valued KDE formulation on three structured output prediction problems; namely, image reconstruction, optical character recognition, and face-to-face mapping. 
%While the output of the first problem is numeric, the second problem has non-numerical output. 
In the first problem, we compare our method using both covariance and conditional covariance operator-valued kernels with the KDE algorithms of~\citet{Weston-KDE-2003} and~\citet{Cortes-KDE-2005}. In the second problem, we evaluate the two implementations of our KDE method with a constrained regression version of KDE~\citep{Cortes-KDE-2007} and  Max-Margin Markov Networks~(M$^3$Ns)~\citep{Taska-2003}. In the third problem, in addition to scalar-valued KDE, we compare them with the joint kernel map (JKM) approach of~\citet{Weston-JKM-2007}.

%%%%%%%%%%%%%%%%%%%%%%%%%%%%%%%%%%%%%%%%%%%%%%%%%%%%%%%%%
%%%%%%%%%%%%%%%%%%%%%%%%%%%%%%%%%%%%%%%%%%%%%%%%%%%%%%%%%
%%%%%%%%%%%%%%%%%%%%%%%%%%%%%%%%%%%%%%%%%%%%%%%%%%%%%%%%%

\begin{table*}[t]
%\vspace{-0.05in} 
\caption{\textbf{(Left)} Performance (mean and standard deviation of RBF loss) of the KDE algorithms of~\citet{Weston-KDE-2003} and~\citet{Cortes-KDE-2005}, and our KDE method with covariance and conditional covariance operator-valued kernels on an {\em image reconstruction problem of handwritten digits}. 
\textbf{(Right)} Performance (mean and standard deviation of Well Recognized word Characters (WRC)) of Max-Margin Markov Networks~(M$^3$Ns) algorithm~\citep{Taska-2003}, constrained regression version of KDE~\citep{Cortes-KDE-2007}, and our  KDE method on an {\em optical character recognition} (OCR) task.} %%with covariance and conditional covariance operator-valued kernels 
%\vspace{-0.05in} 
\centering
\begin{small}
\begin{minipage}[t]{.4\linewidth}
\begin{tabular}{lcccc}
\\[0cm]
\multicolumn{1}{c}{\bf Algorithm}  & $\lambda$ & $\sigma_k$ & $\sigma_l$ & {\bf RBF Loss}  \\ \hline \\[-0.1cm]
KDE - Cortes & $0.01$ & $0.1$ & $10$ &                              $0.9247 \pm 0.0112$  \\
KDE - Weston\protect\footnotemark[2] & $0.01$ & $0.07$ & $10$ &          $0.8145 \pm 0.0131$       \\
KDE - Covariance & $0.1$ & $1$ & $12$ &                           $0.7550 \pm 0.0142$    \\
KDE - Cond. Covariance & $0.1$ &  $1$ & $12$ &             $\bf{0.6276 \pm 0.0106}$    \\[0.15cm]  \hline 
\end{tabular}
\end{minipage}
\hfill
\begin{minipage}[t]{.4\linewidth}
\begin{tabular}{lcc}
\\[0cm]
\multicolumn{1}{c}{\bf Algorithm}  & $\lambda$  & {\bf WRC(\%)}  \\ \hline \\[-0.1cm]
M$^3$Ns & - & $87.0 \pm 0.4$ \\
KDE - Cortes%\protect\footnotemark[3] 
& 0.01 & $88.5 \pm 0.9$ \\
KDE - Covariance & 0.01 & $89.2 \pm 1.5$ \\
KDE - Cond. Covariance & 0.01 &  $\bf{91.8 \pm 1.3}$  \\[0.15cm] \hline
\end{tabular}
\end{minipage}
\end{small}
\label{tableresults2}
%\vspace{-0.15in}
\end{table*}

%\vspace{-0.1in}
\subsection{Image Reconstruction}
\label{ssec:IR}
%\vspace{-0.05in} 

Here we consider the image reconstruction problem used in~\citet{Weston-KDE-2003}. This problem takes the top half~(the first 8 pixel lines) of a USPS postal digit as input and estimates its bottom half. We use exactly the same dataset and setting as in the experiments of~\cite{Weston-KDE-2003}. We apply our KDE method using both covariance and conditional covariance operator-valued kernels and compare it with the KDE algorithms of~\citet{Weston-KDE-2003} and~\citet{Cortes-KDE-2005}. In all these methods, we use RBF kernels for both input and output with the parameters shown in Table~\ref{tableresults2} (left). This table also contains the ridge parameter used by these algorithms. We tried a number of values for these parameters and those in the table yielded the best performance. 

We perform $5$-fold cross validation on the first $1000$ digits of the USPS handwritten $16$ by $16$ pixel digit database, training with a single fold on $200$ examples and testing on the remainder. Given a test input, we solve the problem and then choose as output the pre-image from the training data that is closest to this solution. The loss function used to evaluate the prediction $\hat{y}$ for an output $y$ is the RBF loss induced by the output kernel, i.e.,~$||\Phi_l(y)-\Phi_l(\hat{y})||^2=2-2\exp\big(-||y-\hat{y}||^2/(2\sigma_l^2)\big)$. Table~\ref{tableresults2} (left) shows the mean and standard deviation of the RBF loss for the four KDE algorithms described above. % and the kernel and ridge parameters

Our proposed operator-valued kernel approach showed promising results in this experiment. 
While covariance operator-valued KDE achieved a slight improvement over kPCA-KDE (the algorithm by~\citealt{Weston-KDE-2003}), the conditional covariance operator-valued KDE outperformed all the other algorithms. This improvement in prediction accuracy is due to the fact that the conditional covariance operator allows us to capture the output correlations while taking into account information about the inputs. In this problem, kPCA-based KDE performed better than the KDE formulation of~\citet{Cortes-KDE-2005}. In fact, the latter is equivalent to using an identity-based operator-valued kernel in our formulation, and thus, it is incapable of capturing the dependencies in the output feature space~(contrary to the other methods considered here).

%%%%%%%%%%%%%%%%%%%%%%%%%%%%%%%%%%%%%%%%%%%%%%%%%%%%%%%%%
%%%%%%%%%%%%%%%%%%%%%%%%%%%%%%%%%%%%%%%%%%%%%%%%%%%%%%%%%
%%%%%%%%%%%%%%%%%%%%%%%%%%%%%%%%%%%%%%%%%%%%%%%%%%%%%%%%%

%\vspace{-0.05in}
\subsection{Optical Character Recognition}
\label{ssec:OCR}
%\vspace{-0.05in} 

In order to evaluate the effectiveness of our proposed method in problems with non-numerical outputs, we use an optical character recognition~(OCR) problem. This problem is the one used in~\citet{Taska-2003} and~\citet{Cortes-KDE-2005}. The dataset is a subset of the handwritten words collected by Rob Kassel at the MIT Spoken Language Systems Group. It contains $6,877$  word instances with a total of $52,152$ characters. The image of each character has been normalized into a $16$ by $8$ binary-pixel representation. The OCR task consists in predicting a word from the sequence of pixel-based images of its handwritten characters.  %segmented characters

\footnotetext[2]{These results are obtained using the Spider toolbox available at~www.kyb.mpg.de/bs/people/spider. %We tried different number of Principal Components (PC) from the list \{1,2,3,4,5,10\}, and PC=3 yielded the best performance.
}

Table~\ref{tableresults2} (right) reports the results of our experiments. The performance is measured as the percentage number of word characters correctly recognized (WRC).
We compare our approach with a constrained regression version of Cortes's KDE formulation~\citep{Cortes-KDE-2007} and Max-Margin Markov Networks (M$^3$Ns)~\citep{Taska-2003}. Results for these two methods are reported from~\citep{Cortes-KDE-2007}.
We use exactly the same experimental setup described in~\citep{Cortes-KDE-2007} to evaluate our %covariance and conditional covariance 
operator-valued KDE approach.
%Results with our KDE formulation are obtained using the same experimental setup as the one described in~\cite{Cortes-KDE-2007}. 
%
More precisely, we use \textbf{1)} a $10$-fold cross validation on the $6,877$ words of the dataset, training with a single fold (688 words) and testing on the remainder, \textbf{2)} a polynomial kernel of third degree on the image characters, \textbf{3)} the same input and output feature maps. The feature map associated to an input sequence of images $x=x_1 \ldots x_q$ is defined by $ \Phi_k(x) = \big[ k(c_1,x_{v(c_1)}),\ldots, k(c_N,x_{v(c_N)})\big]^\top$, where $c_m$, $m=1,\ldots, N$, are all the image segments in the training set, $v(c_m)$ is the position of the character $c_m$ in the word, and $k(c_m,x_{v(c_m)}) = 0$ if $v(c_m)>q$. For the output space, the feature map  $\Phi_l(y)$ associated to an output string $y = y_1,\ldots,y_q$ is a $26p$-dimensional vector defined by $\Phi_l(y)= \big[ \phi_l(y_1),\ldots,\phi_l(y_q),0,\ldots,0\big]^\top$, where $p$ is the maximum length of a sequence of images in the training set, and $\phi_l(y_j), 1\leqslant j \leqslant q,$ is a 26-dimensional vector whose 
components are all zero except for the entry of index $y_j$. With this output feature space, the pre-image is easily computed since each position can be obtained separately. Note that this occurs since with the OCR dataset a one-to-one mapping of images to characters is provided.

% \footnotetext[3]{This algorithm is a constrained regression version of KDE proposed in~\citet{Cortes-KDE-2007} that allows us to incorporate prior knowledge on input-output dependencies. %where there is a one-to-one mapping of images to characters in an OCR task.
% }

%
Experiments on the OCR task support the results obtained in the image reconstruction of Sec.~\ref{ssec:IR}. 
While covariance based operator-valued KDE achieved better (but comparable) results than the existing state-of-the-art methods, conditional covariance operator-valued KDE outperformed all the other algorithms. 
%The covariance based operator-valued kernel operates on the output space and encodes the interactions between the outputs without any reference to the input space, while using the conditional covariance we also consider the effects of the input variables, and this leads to performance improvement.

%%%%%%%%%%%%%%%%%%%%%%%%%%%%%%%%%%%%%%%%%%%%%%%%%%%%%%%%%
%%%%%%%%%%%%%%%%%%%%%%%%%%%%%%%%%%%%%%%%%%%%%%%%%%%%%%%%%
%%%%%%%%%%%%%%%%%%%%%%%%%%%%%%%%%%%%%%%%%%%%%%%%%%%%%%%%%

%%%%%%%%%%%%%%%%%%%%%%%%%%%%%%%%%%%%%%%%%%%%%%%%%%%%%%%%%
%%%%%%%%%%%%%%%%%%%%%%%%%%%%%%%%%%%%%%%%%%%%%%%%%%%%%%%%%
%%%%%%%%%%%%%%%%%%%%%%%%%%%%%%%%%%%%%%%%%%%%%%%%%%%%%%%%%

%\vspace{-0.05in}
\subsection{Face-to-Face Mapping}
\label{ssec:FI}
%\vspace{-0.05in} 

In this experiment, we first compare the covariance-based operator-valued KDE with the KDE algorithm of~\citet{Cortes-KDE-2005} and the JKM approach of~\citet{Weston-JKM-2007}, and then show how we can speed up the training of our proposed KDE method using incomplete Cholesky decomposition; see~\citep[Appendix~C]{kadri-2013-ICML_TechRep} for more details on applying incomplete Cholesky decomposition to block kernel matrices associated to separable operator-valued kernels. 
Similar to the ``learning to smile'' experiment in~\citet{Weston-JKM-2007}, we consider the problem of mapping the rotated view of a face to the plain expression (frontal view) of the same face.
For that, we use grey-scale views of human faces taken from the MPI face database\footnote[3]{%The Max Planck Institute (MPI) face database is 
Available at http://faces.kyb.tuebingen.mpg.de.}~\citep{MPIface}. The database contains 
$256\times256$~pixels images of 7 views (frontal and rotated) of 200 laser-scanned heads without hair. % = 65536

\begin{table}[t]
%\vspace{-0.1in} 
\centering
%\begin{small}
\caption{Mean-squared errors (MSE) of the JKM algorithm of~\citet{Weston-JKM-2007}, KDE algorithm of~\citet{Cortes-KDE-2005}, and our KDE method with covariance operator-valued kernels on the face-to-face mapping problem.}
%\vspace{-0.075in} 
\small
\begin{tabular}{lccc}
\\[0cm]
\multicolumn{1}{c}{\bf Algorithm} & $\lambda$ & $\sigma_k$ & {\bf MSE}  \\ \hline \\[-0.1cm]
JKM - Patch kernel & - & 3 & $ 0.1257\pm 0.0016$ \\
KDE - Cortes  & 0.1 & 1 & $0.1773 \pm 0.0012$ \\
KDE - Covariance & 0.1 & 4 & $0.1570 \pm 0.0021$ \\
KDE - Cond. Covariance & 0.1 & 4 &  $\bf{0.1130 \pm 0.0014}$  \\[0.15cm] \hline
\end{tabular}
\label{tableresults3}
\vspace{-0.1in}
%\end{small}
\end{table}

To show the effectiveness of our approach, we use a relatively small number of training examples in our first experiment (similar to~\citealt{Weston-JKM-2007}). We consider the problem of predicting plain expression faces from only 30 degree right rotated views. We use 20 examples for training and 80 for testing. We apply a JKM using the patch-wise joint kernel defined in~\citep{Weston-JKM-2007}, with patches of size $10 \times 10$ that overlap by $5$ pixels. For all the methods (JKM and KDE-based), we use a RBF kernel for inputs and a linear kernel for outputs.
Table~\ref{tableresults3} reports the mean squared error (MSE) obtained by each algorithm. % and the corresponding regularization parameters. 
The results indicate that JKM and conditional covariance KDE algorithms outperform identity and conditional KDE methods, and conditional covariance KDE achieves the best performance. This confirms that we can improve the performance by taking into account the relationship between inputs-outputs.

\begin{figure}[t]
%\vspace{-0.1in} 
\centering
\includegraphics[scale=0.24]{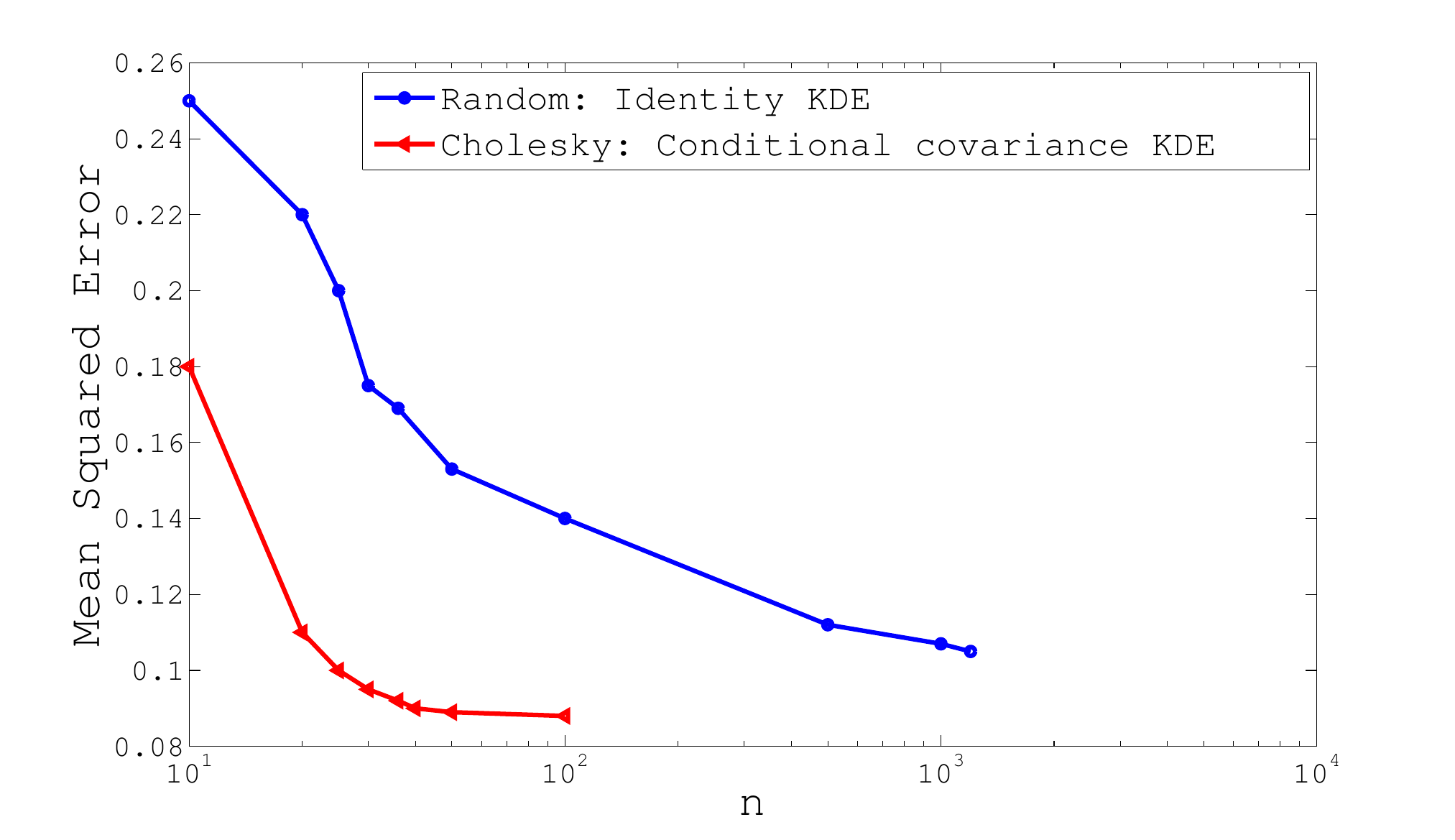}
%\vspace{-0.15in}
\caption{We compare the efficient implementation (using incomplete Cholesky decomposition) of our conditional covariance operator-valued KDE method with the KDE algorithm of~\citet{Cortes-KDE-2005}. While the parameter $n$ is $m_1=m_2=n$ in the incomplete Cholesky decomposition, it is the number of faces randomly selected from 1,200 training faces in the KDE algorithm of~\citet{Cortes-KDE-2005}. The right-most point is the MSE of training on the full training set of $n=1,200$ examples.}
%\vspace{-0.1in}
\label{fig:applicability}
\end{figure}

We now focus on the scalability of our method and consider the face-to-face mapping problem with a large number of examples. Here we use all the rotated face images (30, 60, and 90 degree left and right rotations) to predict the plain face expression. This gives us 1,200 examples for training and 200 for testing. Fig.~\ref{fig:applicability} compares the performance of the efficient implementation (using incomplete Cholesky decomposition) of our conditional covariance operator-valued KDE method with the original KDE algorithm~\citep{Cortes-KDE-2005}. % (i.e.,~identity operator-valued KDE). 
The parameter $n$ is the number of faces randomly selected from 1,200 training faces in the original KDE and is $m_1=m_2=n$ in the incomplete Cholesky decomposition. The results indicate that the low-rank approximation of our KDE method leads to both a considerable reduction in computation time and a good performance. It obtains a better MSE with $m_1 = m_2 = 30$ than the original KDE with all 1,200 examples.

%%%%%%%%%%%%%%%%%%%%%%%%%%%%%%%%%%%%%%%%%%%%%%%%%%%%%%%%%
%%%%%%%%%%%%%%%%%%%%%%%%%%%%%%%%%%%%%%%%%%%%%%%%%%%%%%%%%

%\vspace{-0.1in}
\section{Conclusions and Future Work}
\label{sec:conc}

In this paper, we presented a general formulation of kernel dependency estimation~(KDE) for structured output learning using operator-valued kernels, and illustrated its use in several experiments. 
%Several existing KDE methods including those in~\citep{Weston-KDE-2003,Cortes-KDE-2005} can be recovered from our formulation by choosing a particular operator-valued kernel. 
We also proposed a new covariance-based operator-valued kernel that takes into account the structure of the output kernel feature space. This kernel   encodes the interactions between the outputs, but makes no reference to the input space. We addressed this issue by introducing a variant of our KDE method based on the conditional covariance operator that in addition to the correlation between the outputs takes into account the effects of the input variables. 

% Finally, we evaluated the performance of our KDE method, using both covariance and conditional covariance kernels, on three structured output problems,
%  and compared it to the state-of-the-art kernel-based structured output prediction methods. 

In our work, we focused on regression-based structured output prediction. An interesting direction for future research is to explore operator-valued kernels in the context of classification-based structured output learning. Joint kernels and operator-valued kernels have strong connections, but more investigation is needed to show how operator-valued kernel formulation can be used  to improve joint kernel methods, and how to deal with the pre-image problem in this case.

%%%%%%%%%%%%%%%%%%%%%%%%%%%%%%%%%%%%%%%%%%%%%%%%%%%%%%%%%
%%%%%%%%%%%%%%%%%%%%%%%%%%%%%%%%%%%%%%%%%%%%%%%%%%%%%%%%%
%%%%%%%%%%%%%%%%%%%%%%%%%%%%%%%%%%%%%%%%%%%%%%%%%%%%%%%%%

\section*{Acknowledgments}
 We would like to thank Arthur Gretton and Alain Rakotomamonjy for their
help and discussions. 
This research was funded by the Ministry of Higher Education and Research and
the ANR project LAMPADA~(ANR-09-EMER-007).

%We would like to thank the anonymous reviewers for their valuable comments.

%
% % % This research was funded by the Ministry of Higher Education and Research, 
% % % Nord-Pas-de-Calais Regional Council and FEDER (Contrat de Projets Etat Region CPER 2007-2013), 
% % % the ANR project LAMPADA (ANR-09-EMER-007), 
% % % and by the IST Program of the European Community under the PASCAL2 Network of Excellence (IST-216886). 
%

% This publication only reflects the authors' views.

%%%%%%%%%%%%%%%%%%%%%%%%%%%%%%%%%%%%%%%%%%%%%%%%%%%%%%%%%
%%%%%%%%%%%%%%%%%%%%%%%%%%%%%%%%%%%%%%%%%%%%%%%%%%%%%%%%%
%%%%%%%%%%%%%%%%%%%%%%%%%%%%%%%%%%%%%%%%%%%%%%%%%%%%%%%%%

\newpage
%\begin{small}
\bibliography{Struct-Output}

\begin{thebibliography}{26}
\providecommand{\natexlab}[1]{#1}
\providecommand{\url}[1]{\texttt{#1}}
\expandafter\ifx\csname urlstyle\endcsname\relax
  \providecommand{\doi}[1]{doi: #1}\else
  \providecommand{\doi}{doi: \begingroup \urlstyle{rm}\Url}\fi

\bibitem[{\'A}lvarez et~al.(2012){\'A}lvarez, Rosasco, and
  Lawrence]{Alvarez-2012}
{\'A}lvarez, M., Rosasco, L., and Lawrence, N.~D.
\newblock Kernels for vector-valued functions: a review.
\newblock \emph{Foundations and Trends in Machine Learning}, 4\penalty0
  (3):\penalty0 195--266, 2012.

\bibitem[Bach \& Jordan(2002)Bach and Jordan]{Bach-2002}
Bach, F. and Jordan, M.
\newblock Kernel independent component analysis.
\newblock \emph{Journal of Machine Learning Research}, 3:\penalty0 1--48, 2002.

\bibitem[Bakir et~al.(2007)Bakir, Hofmann, Sch\"{o}lkopf, Smola, Taskar, and
  Vishwanathan]{Bakir-2007}
Bakir, G., Hofmann, T., Sch\"{o}lkopf, B., Smola, A., Taskar, B., and
  Vishwanathan, S. (eds.).
\newblock \emph{Predicting Structured Data}.
\newblock MIT Press, 2007.

\bibitem[Brouard et~al.(2011)Brouard, d{\textquoteright}Alch{\'e} Buc, and
  Szafranski]{Flo-2011}
Brouard, C., d{\textquoteright}Alch{\'e} Buc, F., and Szafranski, M.
\newblock Semi-supervised penalized output kernel regression for link
  prediction.
\newblock In \emph{Proc. ICML}, pp.\  593--600, 2011.

\bibitem[Caponnetto \& {De Vito}(2007)Caponnetto and {De
  Vito}]{Caponnetto-2007}
Caponnetto, A. and {De Vito}, E.
\newblock Optimal rates for the regularized least-squares algorithm.
\newblock \emph{Foundations of Computational Mathematics}, 7\penalty0
  (3):\penalty0 331--368, 2007.

\bibitem[Caponnetto et~al.(2008)Caponnetto, Micchelli, Pontil, and
  Ying]{Caponnetto-2008}
Caponnetto, A., Micchelli, C., Pontil, M., and Ying, Y.
\newblock Universal multi-task kernels.
\newblock \emph{Journal of Machine Learning Research}, 68:\penalty0 1615--1646,
  2008.

\bibitem[Caruana(1997)]{Caruana-1997}
Caruana, R.
\newblock Multitask learning.
\newblock \emph{Machine Learning}, 28\penalty0 (1):\penalty0 41--75, 1997.

\bibitem[Cortes et~al.(2005)Cortes, Mohri, and Weston]{Cortes-KDE-2005}
Cortes, C., Mohri, M., and Weston, J.
\newblock A general regression technique for learning transductions.
\newblock In \emph{Proc. ICML}, pp.\  153--160, 2005.

\bibitem[Cortes et~al.(2007)Cortes, Mohri, and Weston]{Cortes-KDE-2007}
Cortes, C., Mohri, M., and Weston, J.
\newblock \emph{A General Regression Framework for Learning String-to-String
  Mappings}.
\newblock MIT Press, 2007.

\bibitem[Evgeniou et~al.(2005)Evgeniou, Micchelli, and Pontil]{Evgeniou-2005}
Evgeniou, T., Micchelli, C., and Pontil, M.
\newblock Learning multiple tasks with kernel methods.
\newblock \emph{Journal of Machine Learning Research}, 6:\penalty0 615--637,
  2005.

\bibitem[Fukumizu et~al.(2004)Fukumizu, Bach, and Jordan]{Fukumizu-2004}
Fukumizu, K., Bach, F., and Jordan, M.
\newblock Dimensionality reduction for supervised learning with reproducing
  kernel hilbert spaces.
\newblock \emph{Journal of Machine Learning Research}, 5:\penalty0 73--99,
  2004.

\bibitem[Fukumizu et~al.(2011)Fukumizu, Song, and Gretton]{Fukumizu-2011}
Fukumizu, K., Song, L., and Gretton, A.
\newblock Kernel bayes' rule.
\newblock In \emph{Neural Information Processing Systems (NIPS)}, 2011.

\bibitem[Gretton et~al.(2005)Gretton, Herbrich, Smola, Bousquet, and
  Sch\"{o}lkopf]{Gretton-2005}
Gretton, A., Herbrich, R., Smola, A., Bousquet, O., and Sch\"{o}lkopf, B.
\newblock Kernel methods for measuring independence.
\newblock \emph{Journal of Machine Learning Research}, 6:\penalty0 1--47, 2005.

\bibitem[Grunewalder et~al.(2012)Grunewalder, Lever, Gretton, Baldassarre,
  Patterson, and Pontil]{Grunewalder-2011a}
Grunewalder, S., Lever, G., Gretton, A., Baldassarre, L., Patterson, S., and
  Pontil, M.
\newblock Conditional mean embeddings as regressors.
\newblock In \emph{Proc. ICML}, 2012.

\bibitem[Kadri et~al.(2010)Kadri, Duflos, Preux, Canu, and Davy]{Kadri-2010}
Kadri, H., Duflos, E., Preux, P., Canu, S., and Davy, M.
\newblock Nonlinear functional regression: a functional {RKHS} approach.
\newblock In \emph{Proc. AISTATS}, pp.\  111--125, 2010.

\bibitem[Kadri et~al.(2011)Kadri, Rabaoui, Preux, Duflos, and
  Rakotomamonjy]{Kadri-ICML-2011}
Kadri, H., Rabaoui, A., Preux, P., Duflos, E., and Rakotomamonjy, A.
\newblock Functional regularized least squares classification with
  operator-valued kernels.
\newblock In \emph{Proc. ICML}, pp.\  993--1000, 2011.

\bibitem[Kadri et~al.(2012)Kadri, Ghavamzadeh, and
  Preux]{kadri-2013-ICML_TechRep}
Kadri, H., Ghavamzadeh, M., and Preux, P.
\newblock A generalized kernel approach to structured output learning.
\newblock Technical Report 00695631, INRIA, 2012.

\bibitem[Micchelli \& Pontil(2005)Micchelli and Pontil]{Micchelli-2005a}
Micchelli, C. and Pontil, M.
\newblock On learning vector-valued functions.
\newblock \emph{Neural Computation}, 17:\penalty0 177--204, 2005.

\bibitem[Ramsay \& Silverman(2005)Ramsay and Silverman]{Ramsay-2005}
Ramsay, J. and Silverman, B.
\newblock \emph{Functional Data Analysis, 2nd edition}.
\newblock Springer Verlag, New York, 2005.

\bibitem[Sch\"{o}lkopf et~al.(1999)Sch\"{o}lkopf, Mika, Burges, Knirsch,
  M\"{u}ller, R\"{a}tsch, and Smola]{Scholkopf-1999}
Sch\"{o}lkopf, B., Mika, S., Burges, C. J.~C., Knirsch, P., M\"{u}ller, K.-R.,
  R\"{a}tsch, G., and Smola, A.~J.
\newblock Input space vs.\ feature space in kernel-based methods.
\newblock \emph{IEEE Trans. on Neural Networks}, 10\penalty0 (5):\penalty0
  1000--1017, 1999.

\bibitem[Taskar et~al.(2004)Taskar, Guestrin, and Koller]{Taska-2003}
Taskar, B., Guestrin, C., and Koller, D.
\newblock Max-margin markov networks.
\newblock In \emph{Advances in Neural Information Processing Systems 16}. 2004.

\bibitem[Troje \& Bulthoff(1996)Troje and Bulthoff]{MPIface}
Troje, N. and Bulthoff, H.
\newblock Face recognition under varying poses: The role of texture and shape.
\newblock \emph{Vision Research}, 36:\penalty0 1761--1771, 1996.

\bibitem[Tsochantaridis et~al.(2005)Tsochantaridis, Joachims, Hofmann, and
  Altun]{Tsochantaridis-2005}
Tsochantaridis, I., Joachims, T., Hofmann, T., and Altun, Y.
\newblock Large margin methods for structured and interdependent output
  variables.
\newblock \emph{Journal of machine Learning Research}, 6:\penalty0 1453--1484,
  2005.

\bibitem[Wang \& Shawe-Taylor(2010)Wang and Shawe-Taylor]{Wang-2010}
Wang, Z. and Shawe-Taylor, J.
\newblock A kernel regression framework for {SMT}.
\newblock \emph{Machine Translation}, 24\penalty0 (2):\penalty0 87--102, 2010.

\bibitem[Weston et~al.(2003)Weston, Chapelle, Elisseeff, Sch\"{o}lkopf, and
  Vapnik]{Weston-KDE-2003}
Weston, J., Chapelle, O., Elisseeff, A., Sch\"{o}lkopf, B., and Vapnik, V.
\newblock Kernel dependency estimation.
\newblock In \emph{Proceedings of the Advances in Neural Information Processing
  Systems 15}, pp.\  873--880, 2003.

\bibitem[Weston et~al.(2007)Weston, BakIr, Bousquet, Sch\"{o}lkopf, Mann, and
  Noble]{Weston-JKM-2007}
Weston, J., BakIr, G., Bousquet, O., Sch\"{o}lkopf, B., Mann, T., and Noble, W.
\newblock \emph{Joint Kernel Maps}.
\newblock MIT Press, 2007.

\end{thebibliography}
\bibliographystyle{icml2013}
%\end{small}

%%%%%%%%%%%%%%%%%%%%%%%%%%%%%%%%%%%%%%%%%%%%%%%%%%%%%%%%%
%%%%%%%%%%%%%%%%%%%%%%%%%%%%%%%%%%%%%%%%%%%%%%%%%%%%%%%%%
%%%%%%%%%%%%%%%%%%%%%%%%%%%%%%%%%%%%%%%%%%%%%%%%%%%%%%%%%
%%%%%%%%%%%%%%%%%%%%%%%%%%%%%%%%%%%%%%%%%%%%%%%%%%%%%%%%%
%%%%%%%%%%%%%%%%%%%%%%%%%%%%%%%%%%%%%%%%%%%%%%%%%%%%%%%%%

\newpage
\onecolumn

\section{Appendix}

\subsection{(Generalized) Kernel Trick} 
\label{appxc1}

In this section, we prove the (generalized) kernel trick used in Section~\ref{sec:kde} of the paper, i.e., $\langle \mathcal{T} \Phi_l(y_1),\Phi_l(y_2)\rangle_\mathcal{F_Y} = [\mathcal{T}l(y_1,\cdot)](y_2)$, 
where $\mathcal{T} \in \mathcal{L(F_Y)}$ and $\mathcal{F_Y}$ is a RKHS with kernel $l$. \\

\noindent
\textbf{Case~1:} $\Phi_l$ is the feature map associated to the reproducing kernel $l$, i.e.,~$\Phi_l(y) = l(\cdot,y)$. \\

Here the proof is straightforward, we may write
\begin{equation*}
\langle \mathcal{T} \Phi_l(y_1),\Phi_l(y_2)\rangle_\mathcal{F_Y} = \langle \mathcal{T} l(\cdot, y_1),l(\cdot,y_2)\rangle_\mathcal{F_Y} = [\mathcal{T}l(y_1,\cdot)](y_2).  
\end{equation*}
The second equality follows from the reproducing property. \\

\noindent
\textbf{Case~2:} $\Phi_l$ is an implicit feature map of a Mercer kernel, and $\mathcal{T}$ is a self-adjoint operator in $\mathcal{L(F_Y)}$. \\
 
We first recall the Mercer's theorem:
\begin{theo}[Mercer's theorem] Suppose that $l$ is a symmetric real-valued kernel on $\mathcal{Y}^2$ such that the integral operator $T_l: L_2(\mathcal{Y}) \rightarrow L_2(\mathcal{Y})$, defined as
\begin{equation*}
 (T_l f)(y_1) := \int_\mathcal{Y} l(y_1,y_2) f(y_2) dy_2
\end{equation*}
is positive. Let $\gamma_j \in L_2(\mathcal{Y})$ be the normalized eigenfunctions of $T_l$ associated with the 
eigenvalues $\lambda_j > 0$, sorted in non-increasing order. Then
\begin{equation}
\label{eq:app1}
l(y_1,y_2) = \sum_{j=1}^{N_f} \lambda_j \gamma_j(y_1)  \gamma_j(y_2) 
\end{equation}
holds for almost all $(y_1,y_2)$ with $N_f \in \mathbb{N}$.
\end{theo}
Since $l$ is a Mercer kernel, the eigenfunctions $(\gamma_i)_{i=1}^{N_f}$ can be chosen to be orthogonal w.r.t.~the dot product in $L_2(\mathcal{Y})$. Hence, it is straightforward to construct a dot product $\langle \cdot, \cdot \rangle$ such that 
$\langle \gamma_i, \gamma_j \rangle = \delta_{ij} / \lambda_j$ ($\delta_{ij}$ is the Kronecker delta) and the orthonormal basis $\big(e_j\big)_{j=1}^{N_f} = \big(\sqrt{\lambda_j} \gamma_j\big)_{j=1}^{N_f}$ (see~\citep{Scholkopf-1999} for more details). Therefore, the feature map associated to the Mercer kernel $l$ is of the form
\begin{equation*}
 \Phi_l: y \longmapsto \big(\sqrt{\lambda_j} \gamma_j(y) \big)_{j=1}^{N_f}\;.
\end{equation*}
Using~\eqref{eq:app1}, we can compute $[\mathcal{T}l(y_1,\cdot)](y_2)$ as follows: 
\begin{align}
\label{eq:app2}
 [\mathcal{T}l(y_1,\cdot)](y_2) & =  \sum_{j=1}^{N_f} \lambda_j \gamma_j(y_1)\big[\mathcal{T} \gamma_j\big](y_2) \nonumber \\
 & =  \sum_{i,j=1}^{N_f} \lambda_j \gamma_j(y_1) \langle \mathcal{T} \gamma_j , e_i \rangle e_i(y_2) \nonumber \\
& = \  \sum_{i,j=1}^{N_f} \lambda_j \gamma_j(y_1) \lambda_i \langle \mathcal{T} \gamma_j , \gamma_i \rangle \gamma_i(y_2).
\end{align}
Let $\widehat{\mathcal{T}} = \big(\widehat{\mathcal{T}}_{ij}\big)_{i,j=1}^{N_f}$ be the matrix representation of the operator $\mathcal{T}$ in the basis $\big(e_j\big)_{j=1}^{N_f}$. By definition we have $\widehat{\mathcal{T}}_{ij} = \langle \mathcal{T} e_i, e_j \rangle$. Using this and the feature map expression of a Mercer kernel, we obtain
\begin{align}
\label{eq:app3}
\langle \mathcal{T} \Phi_l(y_1)&,\Phi_l(y_2)\rangle_\mathcal{F_Y} = \sum_{i=1}^{N_f} \big(\widehat{\mathcal{T}} \Phi_l(y_1)\big)_i \big(\Phi_l(y_2)\big)_i \nonumber \\
& = \sum_{i=1}^{N_f} \left(\sum_{j=1}^{N_f}\widehat{\mathcal{T}}_{ij} \sqrt{\lambda_j}  \gamma_j(y_1)\right) \sqrt{\lambda_i} \gamma_i(y_2) \nonumber \\
& =  \sum_{i,j=1}^{N_f} \langle \mathcal{T} e_i, e_j \rangle \sqrt{\lambda_j}  \gamma_j(y_1) \sqrt{\lambda_i} \gamma_i(y_2) \nonumber \\
& =  \sum_{i,j=1}^{N_f} \langle \mathcal{T} \sqrt{\lambda_i}\gamma_i, \sqrt{\lambda_j}\gamma_j \rangle \sqrt{\lambda_j}  \gamma_j(y_1) \sqrt{\lambda_i} \gamma_i(y_2) \nonumber \\
& =  \sum_{i,j=1}^{N_f} \lambda_j \gamma_j(y_1) \langle \mathcal{T} \gamma_i, \gamma_j \rangle  \lambda_i \gamma_i(y_2) \nonumber \\
& =  \sum_{i,j=1}^{N_f} \lambda_j \gamma_j(y_1) \langle \mathcal{T} \gamma_j, \gamma_i \rangle  \lambda_i \gamma_i(y_2). 
\end{align}

Note that the last equality follows from the fact that $\mathcal{T}$ is a self-adjoint operator. The proof follows from Eqs.~\ref{eq:app2} and~\ref{eq:app3}.
%%
%\begin{center}
% (i) and (ii) $ \Rightarrow  \langle \mathcal{T} \Phi_l(y_1),\Phi_l(y_2)\rangle_\mathcal{F_Y} = [\mathcal{T}l(y_1,\cdot)](y_2)$. 
%\end{center}

%%%%%%%%%%%%%%%%%%%%%%%%%%%%%%%%%%%%%%%%%%%%%%%%%%%%%%%%%
%%%%%%%%%%%%%%%%%%%%%%%%%%%%%%%%%%%%%%%%%%%%%%%%%%%%%%%%%
%%%%%%%%%%%%%%%%%%%%%%%%%%%%%%%%%%%%%%%%%%%%%%%%%%%%%%%%%

\subsection{Proof of Proposition 1}
\label{appx2}

In this section, we provide the proof of Proposition~\ref{prop:cov}. We only show the proof for the covariance-based operator-valued kernels, since the proof for the other case (conditional covariance-based operator-valued kernels) is quite similar. Note that the pre-image problem is of the form
\begin{equation*}
 f(x) = \argmin\limits_{y\in \mathcal{Y}}\;l(y,y) - 2 \big[\mathbf{K}_x (\mathbf{K} + \lambda I)^{-1} \mathbf{L}_\bullet\big](y),
%\sum\limits_i [(K_x (\mathbf{K} + \lambda I)^{-1})_i l(y_i,.)](y)
\end{equation*}
and our goal is to compute its Gram matrix expression\footnote[4]{Expressing the covariance operator $\widehat{C}_{YY}^{(n)}$ on the RKHS~$\mathcal{F_Y}$ using the kernel matrix $\mathbf{L}$.} in case $K(x_i,x_j) = k(x_i,x_j) \widehat{C}_{YY}^{(n)}$, where $\widehat{C}_{YY}^{(n)}$ is the empirical covariance operator defined as
\begin{equation*}
\widehat{C}_{YY}^{(n)}=\frac{1}{n}\sum\limits_{i=1}^n l(\cdot,y_i)\otimes l(\cdot,y_i).
\end{equation*}

Let $h = (h_i)_{i=1}^n$ be a vector of variables in the RKHS $\mathcal{F_Y}$ such that $h = (\mathbf{K} + \lambda I)^{-1} \mathbf{L}_\bullet$. Since each $h_i$ is in the RKHS $\mathcal{F_Y}$, it can be decomposed as 
\begin{equation*}
h_i = \alpha_{(i)}^{\top} \mathbf{L}_\bullet + h_{i\bot} = \sum\limits_{j=1}^n \alpha_{(i)}^j l(\cdot,y_j) + h_{i\bot}, 
\end{equation*}
where $\alpha_{(i)}\in \mathbb{R}^n$, $\mathbf{L}_\bullet = \big(l(\cdot,y_1),\ldots,l(\cdot,y_n)\big)^\top$, and $h_{i\bot}$ is orthogonal to all $l(\cdot,y_i)$'s, $i=1,\ldots,n$. The idea here is similar to the one used by~\citet{Fukumizu-2011}. Now we may write
\begin{equation*}
(\mathbf{K} + \lambda I) h = \mathbf{L}_\bullet, 
\end{equation*} 
which gives us
\begin{equation*} 
\forall i \in {1,\ldots,n} \quad\quad l(\cdot,y_i)=\sum_{j=1}^n \mathbf{K}_{ij} h_j + \lambda h_i.
\end{equation*}
Using the empirical covariance operator, for each $i$, we may write
\begin{align*}
l(\cdot,y_i)&=\sum_{j=1}^nk(x_i,x_j) \widehat{C}_{YY}^{(n)}h_j+\lambda h_i \\
&=\sum_{j=1}^nk(x_i,x_j)\Big(\frac{1}{n}\sum_{s=1}^nl(\cdot,y_s)\otimes l(\cdot,y_s)\Big)h_j+\lambda h_i \\
&=\sum_{j=1}^n\frac{1}{n}k(x_i,x_j)\Big(\sum_{s=1}^nl(\cdot,y_s)\otimes l(\cdot,y_s)\Big) \Big(\sum_{m=1}^n\alpha_{(j)}^ml(\cdot,y_m)+h_{i\bot}\Big)+\lambda\sum_{m=1}^n\alpha_{(i)}^ml(\cdot,y_m)+\lambda h_{i\bot} \\
&=\sum_{j=1}^n\frac{1}{n}k(x_i,x_j)\sum_{s=1}^n\sum_{m=1}^n\alpha_{(j)}^ml(y_s,y_m)l(\cdot,y_s)+0+\lambda\mathbf{L}_\bullet^\top\alpha_{(i)}+\lambda h_{i\bot} \\
&=\sum_{j=1}^n\frac{1}{n}k(x_i,x_j)\sum_{s=1}^nl(\cdot,y_s)\sum_{m=1}^n\alpha_{(j)}^ml(y_s,y_m)+\lambda\mathbf{L}_\bullet^\top\alpha_{(i)}+\lambda h_{i\bot} \\
&=\sum_{j=1}^n\frac{1}{n}k(x_i,x_j)\mathbf{L}_\bullet^\top\mathbf{L}\alpha_{(j)}+\lambda\mathbf{L}_\bullet^\top \alpha_{(i)} + \lambda h_{i\bot}.
\end{align*}
Now if take the inner-product of the above equation with all $l(y_s,\cdot),\;s=1,\ldots,n$, we obtain that for each $i$
\begin{align*}
l(y_i,y_s)&=\sum_{j=1}^n\frac{1}{n}k(x_i,x_j)\langle l(\cdot,y_s),\mathbf{L}_\bullet^\top\mathbf{L}\alpha_{(j)}\rangle +\lambda \langle l(\cdot,y_s),\mathbf{L}_\bullet^\top\alpha_{(i)}\rangle+\lambda \langle l(\cdot,y_s),h_{i\bot}\rangle \\
&=\sum_{j=1}^n\frac{1}{n}k(x_i,x_j)\mathbf{L}_{y_s}^\top\mathbf{L}\alpha_{(j)}+\lambda\mathbf{L}_{y_s}^\top\alpha_{(i)},
\end{align*}
which gives us the vector form
\begin{equation}
\label{eq:app4}
\mathbf{L}_{y_i}=\sum_{j=1}^n \frac{1}{n} k(x_i,x_j) \mathbf{L} \mathbf{L} \alpha_{(j)} + \lambda  \mathbf{L} \alpha_{(i)}. 
\end{equation}
Defining the $n\times n$ matrix $\boldsymbol{\alpha}=(\alpha_{(1)},\ldots,\alpha_{(n)})$, we may write Eq.~\ref{eq:app4} in a matrix form as 
\begin{equation*}
 \mathbf{L}=\frac{1}{n} \mathbf{L} \mathbf{L}\boldsymbol{\alpha}\mathbf{k} + \lambda  \mathbf{L}\boldsymbol{\alpha},
\end{equation*}
which gives us
\begin{equation*}
\frac{1}{n} \mathbf{L}\boldsymbol{\alpha}\mathbf{k} + \lambda\boldsymbol{\alpha} =  I_n.
\end{equation*}
Using $\veco(ABC) = (C^\top \otimes A) \veco(B)$, we have
\begin{equation}
\label{eq:app5}
\big( \frac{1}{n} \mathbf{k} \otimes \mathbf{L} + \lambda I_{n^2} \big) \veco(\boldsymbol{\alpha}) = \veco(I_n).
\end{equation}
Now we may write
\begin{align*}
\mathbf{K}_x (\mathbf{K} + \lambda I)^{-1} \mathbf{L}_\bullet \ &= \ \mathbf{K}_x h = \sum_{i=1}^n K(x,x_i)h_i \\
&= \sum_{i=1}^n k(x,x_i) \widehat{C}_{YY}^{(n)} h_i =\sum_{i=1}^n\frac{1}{n}k(x,x_i)\mathbf{L}_\bullet^\top\mathbf{L}\alpha_{(i)} \\
&= \frac{1}{n} \mathbf{L}_\bullet^\top \mathbf{L}\boldsymbol{\alpha}\mathbf{k}_x = \frac{1}{n} \mathbf{L}_\bullet^\top \veco(\mathbf{L}\boldsymbol{\alpha}\mathbf{k}_x) = \frac{1}{n} \mathbf{L}_\bullet^\top \big(\mathbf{k}^\top_x\otimes \mathbf{L} \big) \veco(\boldsymbol{\alpha}) \\
& = \mathbf{L}_\bullet^\top \big(\mathbf{k}^\top_x\otimes \mathbf{L} \big)  
\big( \mathbf{k} \otimes \mathbf{L} + n \lambda I_{n^2} \big)^{-1} \veco(I_n),
\end{align*}
where the last equality comes from~\eqref{eq:app5}. Thus, we may write
\begin{equation*}
 f(x)  \\ %= \argmin\limits_{y\in \mathcal{Y}}\;l(y,y) - 2 \big[\mathbf{K}_x (\mathbf{K} + \lambda I)^{-1} \mathbf{L}_\bullet\big](y) \\
 =  \argmin\limits_{y\in \mathcal{Y}}\;l(y,y) - 2 \mathbf{L}_y^\top (\mathbf{k}^\top_x\otimes \mathbf{L}) (\mathbf{k} \otimes \mathbf{L} + n \lambda I_{n^2})^{-1} \veco(I_{n}),
\end{equation*}
which concludes the proof.

%%%%%%%%%%%%%%%%%%%%%%%%%%%%%%%%%%%%%%%%%%%%%%%%%%%%%%%%%
%%%%%%%%%%%%%%%%%%%%%%%%%%%%%%%%%%%%%%%%%%%%%%%%%%%%%%%%%
%%%%%%%%%%%%%%%%%%%%%%%%%%%%%%%%%%%%%%%%%%%%%%%%%%%%%%%%%

\subsection{Computational Complexity of Solving the Pre-image Problem}
\label{appx3}

As discussed in Section~\ref{sec:cov}, solving the pre-image problem of Eq.~\ref{eq:prop1} requires computing the following expression:
\begin{equation}
\label{eq:app6}
 C(x,y)  \\ %= \argmin\limits_{y\in \mathcal{Y}}\;l(y,y) - 2 \big[\mathbf{K}_x (\mathbf{K} + \lambda I)^{-1} \mathbf{L}_\bullet\big](y) \\
 =  l(y,y) - 2 \mathbf{L}_y^\top (\mathbf{k}^\top_x\otimes \mathbf{L}) (\mathbf{k} \otimes \mathbf{L} + n \lambda I_{n^2})^{-1} \veco(I_{n}).
\end{equation}
Simple computation of $C(x,y)$ requires storing and inverting the matrix $(\mathbf{k} \otimes \mathbf{L} + n \lambda I_{n^2})\in \mathbb{R}^{n^2\times n^2}$. This leads to space and computational complexities of order $O(n^4)$ and $O(n^6)$. In this section, we provide an efficient procedure to for this computation that reduces its space and computational complexities to $O\big(\max(nm_1m_2\;,\;m_1^2m_2^2)\big)$ and $O(m_1^3m_2^3)$.

We first apply incomplete Cholesky decomposition to the kernel matrices $\mathbf{k} \in \mathbb{R}^{n \times n}$ and $\mathbf{L} \in \mathbb{R}^{n\times n}$~\citep{Bach-2002}. This consists of finding the matrices $\mathbf{U} \in \mathbb{R}^{n \times m_1}$ and $\mathbf{V} \in \mathbb{R}^{n \times m_2}$, with $m_1 \ll n$ and $m_2 \ll n$, such that
\begin{equation*}
\mathbf{k} = \mathbf{U}\mathbf{U}^\top \quad\quad\quad,\quad\quad\quad \mathbf{L} = \mathbf{V}\mathbf{V}^\top.
\end{equation*}
Using this decomposition in~\eqref{eq:app6}, we obtain

\begin{small}
\begin{align}
\label{eq:app7}
 C(x,y) & = l(y,y) - 2 \mathbf{L}_y^\top (\mathbf{k}^\top_x\otimes \mathbf{L}) [\mathbf{U}\mathbf{U}^\top \otimes \mathbf{V}\mathbf{V}^\top + n \lambda I_{n^2}]^{-1} \veco(I_{n}) \nonumber \\
& \stackrel{(a)}{=} l(y,y) - 2 \mathbf{L}_y^\top (\mathbf{k}^\top_x\otimes \mathbf{L}) \big[(\mathbf{U} \otimes \mathbf{V}) (\mathbf{U}^\top \otimes \mathbf{V}^\top) + n \lambda I_{n^2}\big]^{-1} \veco(I_{n}) \nonumber \\
& \stackrel{(b)}{=} l(y,y) - \frac{2}{n\lambda} \mathbf{L}_y^\top (\mathbf{k}^\top_x\otimes \mathbf{L})\Big[I_{n^2} - (\mathbf{U} \otimes \mathbf{V})\big(n\lambda I_{m_1 m_2} + (\mathbf{U}^\top \otimes \mathbf{V}^\top) (\mathbf{U} \otimes \mathbf{V})\big)^{-1} (\mathbf{U}^\top \otimes \mathbf{V}^\top)\Big]  \veco(I_{n}) \nonumber \\
& \stackrel{(c)}{=} l(y,y) -  \frac{2}{n\lambda} \mathbf{L}_y^\top (\mathbf{k}^\top_x\otimes \mathbf{L}) \big[\veco(I_{n}) - (\mathbf{U} \otimes \mathbf{V}) (n\lambda I_{m_1 m_2} + \mathbf{U}^\top \mathbf{U} \otimes \mathbf{V}^\top \mathbf{V})^{-1} (\mathbf{U}^\top \otimes \mathbf{V}^\top)  \veco(I_{n})\big] \nonumber \\
& = l(y,y) -  \frac{2}{n\lambda} \mathbf{L}_y^\top \big[(\mathbf{k}^\top_x\otimes \mathbf{L}) \veco(I_{n}) - (\mathbf{k}^\top_x\otimes \mathbf{L})  (\mathbf{U} \otimes \mathbf{V}) (n\lambda I_{m_1 m_2} + \mathbf{U}^\top \mathbf{U} \otimes \mathbf{V}^\top \mathbf{V})^{-1} (\mathbf{U}^\top \otimes \mathbf{V}^\top)  \veco(I_{n})\big] \nonumber \\
& \stackrel{(d)}{=} l(y,y) -  \frac{2}{n\lambda} \mathbf{L}_y^\top \big[\veco(\mathbf{L}\mathbf{k}_x) - (\mathbf{k}^\top_x \mathbf{U} \otimes \mathbf{L} \mathbf{V}) (n\lambda I_{m_1 m_2} + \mathbf{U}^\top \mathbf{U} \otimes \mathbf{V}^\top \mathbf{V})^{-1} \veco(\mathbf{V}^\top\mathbf{U})\big] \nonumber \\
& \stackrel{(e)}{=} l(y,y) -  \frac{2}{n\lambda} \mathbf{L}_y^\top \big[\mathbf{L}\mathbf{k}_x - (\mathbf{k}^\top_x \mathbf{U} \otimes \mathbf{L} \mathbf{V}) (n\lambda I_{m_1 m_2} + \mathbf{U}^\top \mathbf{U} \otimes \mathbf{V}^\top \mathbf{V})^{-1} \veco(\mathbf{V}^\top\mathbf{U})\big]
\end{align}
\end{small}

{\bf (a)} and {\bf (c)} follow from the fact that $\;\;\;(\mathbf{A}\otimes \mathbf{B})(\mathbf{C}\otimes \mathbf{D})=\mathbf{AC} \otimes \mathbf{BD}$. 

{\bf (b)} follows from the Woodbury formula, i.e.,~$(\mathbf{A}+ \mathbf{BC})^{-1} = \mathbf{A}^{-1} - \mathbf{A}^{-1}\mathbf{B}(\mathbf{I} + \mathbf{CA}^{-1}\mathbf{B})^{-1}\mathbf{CA}^{-1}$.

{\bf (d)} follows from the fact that $\;\;\;\veco(\mathbf{ABC}) = (\mathbf{C}^\top \otimes \mathbf{A})\veco(\mathbf{B})$.

{\bf (e)} follows from the fact that $\mathbf{L}\mathbf{k}_x$ is a $n\times 1$ vector. \\

The most expensive computations in Eq.~\ref{eq:app7} is the inversion of matrix $(n\lambda I_{m_1 m_2} +\mathbf{U}^\top \mathbf{U} \otimes \mathbf{V}^\top \mathbf{V})\in\mathbb{R}^{m_1m_2 \times m_1m_2}$, with computational cost of order $O(m_1^3m_2^3)$. Therefore, we have reduced the cost of computing $C(x,y)$ from $O(n^6)$ to $O(m_1^3m_2^3)$. Moreover, the largest size matrix that is needed to be stored in order to compute Eq.~\ref{eq:app7} is either $(n\lambda I_{m_1 m_2} + \mathbf{U}^\top \mathbf{U} \otimes \mathbf{V}^\top \mathbf{V})\in\mathbb{R}^{m_1m_2\times m_1m_2}$ or $(\mathbf{k}^\top_x \mathbf{U} \otimes \mathbf{L} \mathbf{V})\in\mathbb{R}^{n\times m_1m_2}$, which reduces the space complexity from $O(n^4)$ to $O\big(\max(nm_1m_2\;,\;m_1^2m_2^2)\big)$.

%%%%%%%%%%%%%%%%%%%%%%%%%%%%%%%%%%%%%%%%%%%%%%%%%%%%%%%%%
%%%%%%%%%%%%%%%%%%%%%%%%%%%%%%%%%%%%%%%%%%%%%%%%%%%%%%%%%
%%%%%%%%%%%%%%%%%%%%%%%%%%%%%%%%%%%%%%%%%%%%%%%%%%%%%%%%%
\newpage

\begin{figure}
\centering
 \includegraphics[scale=0.43]{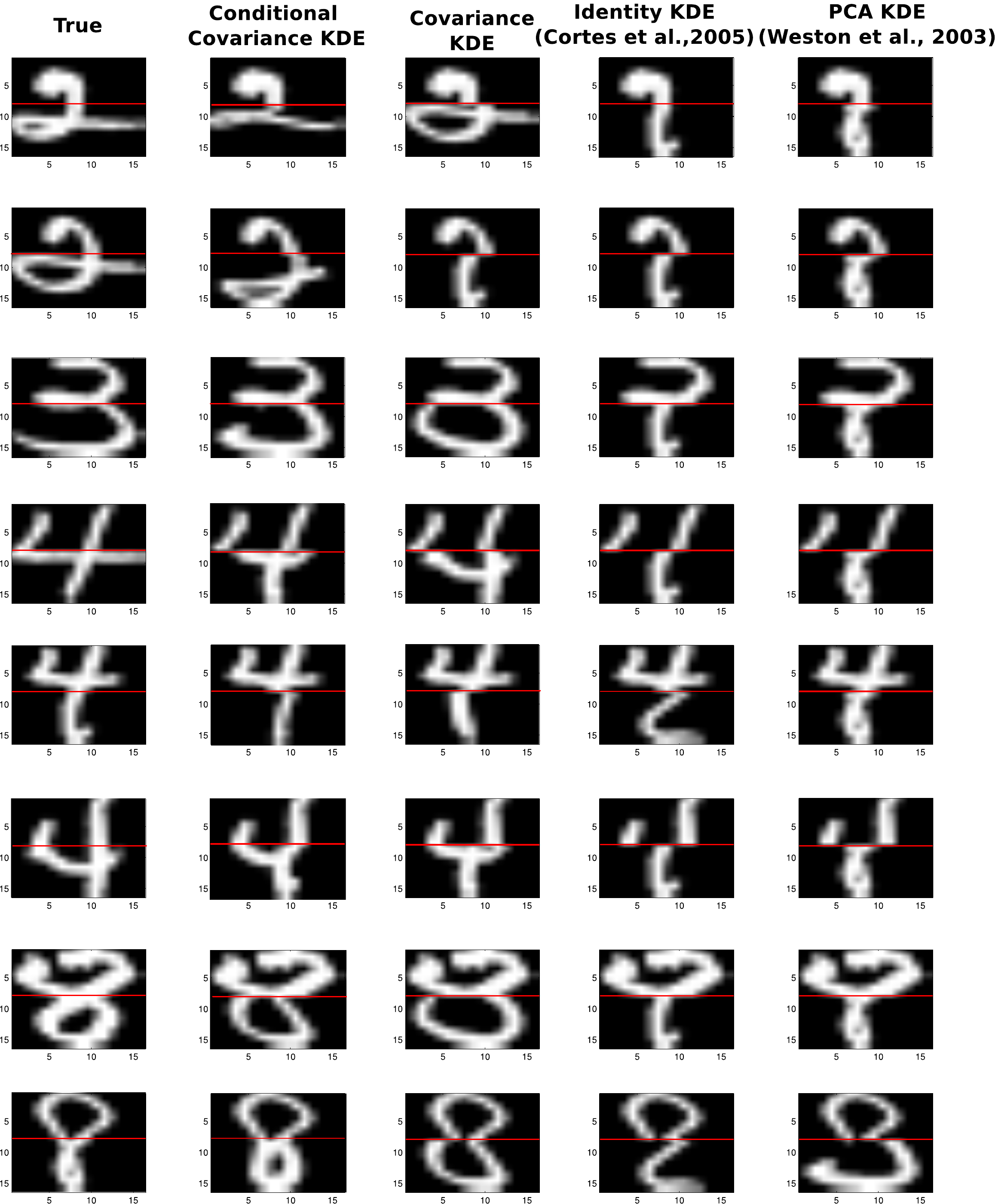}
\end{figure}

%%%%%%%%%%%%%%%%%%%%%%%%%%%%%%%%%%%%%%%%%%%%%%%%%%%%%%%%%
%%%%%%%%%%%%%%%%%%%%%%%%%%%%%%%%%%%%%%%%%%%%%%%%%%%%%%%%%
%%%%%%%%%%%%%%%%%%%%%%%%%%%%%%%%%%%%%%%%%%%%%%%%%%%%%%%%%

\end{document}